\documentclass[sn-mathphys-num]{sn-jnl}

\usepackage{graphicx}%
\usepackage{multirow}%
\usepackage{amsmath,amssymb,amsfonts}%
\usepackage{amsthm}%
\usepackage{mathrsfs}%
\usepackage[title]{appendix}%
\usepackage{xcolor}%
\usepackage{textcomp}%
\usepackage{manyfoot}%
\usepackage{booktabs}%
\usepackage{algorithm}%
\usepackage{algorithmicx}%
\usepackage{algpseudocode}%
\usepackage{listings}%
\usepackage{makecell}
\usepackage{array}
\usepackage{stfloats}
\usepackage{url}
\usepackage{verbatim}
\usepackage{marvosym}
\usepackage{setspace}

\theoremstyle{thmstyleone}%
\theoremstyle{thmstyletwo}%

\theoremstyle{thmstylethree}%

\begin{document}


\title[Article Title]{DefFiller: Mask-Conditioned Diffusion for Salient Steel Surface Defect Generation}


\author[1]{\fnm{Yichun} \sur{Tai}}

\author[1]{\fnm{Zhenzhen} \sur{Huang}}

\author[1]{\fnm{Tao} \sur{Peng}}

\author*[1]{\fnm{Zhijiang} \sur{Zhang}}


\affil[1]{\orgdiv{School of Communication and Information Engineering}, \orgname{Shanghai University}, \orgaddress{\street{Shangda Road 99}, \city{Shanghai}, \postcode{200444}, \state{Shanghai}, \country{China}}}

\abstract{Current saliency-based defect detection methods show promise in industrial settings, but the unpredictability of defects in steel production environments complicates dataset creation, hampering model performance. Existing data augmentation approaches using generative models often require pixel-level annotations, which are time-consuming and resource-intensive. To address this, we introduce DefFiller, a mask-conditioned defect generation method that leverages a layout-to-image diffusion model. DefFiller generates defect samples paired with mask conditions, eliminating the need for pixel-level annotations and enabling direct use in model training. We also develop an evaluation framework to assess the quality of generated samples and their impact on detection performance. Experimental results on the SD-Saliency-900 dataset demonstrate that DefFiller produces high-quality defect images that accurately match the provided mask conditions, significantly enhancing the performance of saliency-based defect detection models trained on the augmented dataset. The code is available at: \url{https://github.com/CC-T/DefFiller}.}




\maketitle

\section{Introduction}
\label{sec1}

Steel is a core material in industrial manufacturing, and detecting surface defects is crucial for enhancing its appearance and reliability~\cite{wu2024efficient}. 
The rapid advancement of deep learning technologies has significantly improved detection capabilities~\cite{zhang2023transformer,tian2023ilidviz}. 
Various saliency-based defect detection models~\cite{sun2024adversarial,ding2022cross,han2022two,shen2024minet,wan2023lfrnet} have been implemented. However, the unpredictable nature of defects in the complex steel production environment complicates the construction of datasets. This can make it more difficult for the detection models to work effectively.

To address this issue, the most straightforward approach is to expand the dataset by generating defect samples. For instance, Wei~\textit{et al.}~\cite{wei2023diversified} develop DCDGANc to replicate real defects, which can then be combined with defect-free images using an enhanced Poisson blending technique. Duan~\textit{et al.}~\cite{Duan2023DFMGAN} introduce DFMGAN to create defective images and masks using their proposed defect-aware residual blocks. However, these methods require training models from scratch and often struggle to precisely align the generated defects with the defective pixels in the images~\cite{li2024novel}.

Recently, diffusion models~\cite{ho2020denoising,rombach2022high,balaji2022ediffi} have gained popularity for sample generation in data augmentation due to their impressive generative capabilities. Wu~\textit{et al.}~\cite{wu2024ddfa} develop a displacement and diffusion-based feature augmentation method to create diverse and high-fidelity samples for tail classes. Similarly, Yang~\textit{et al.}~\cite{yang2024novel} proposed training a Denoising Diffusion Probabilistic Model (DDPM)~\cite{ho2020denoising} to expand fault diagnosis datasets, while Tai~\textit{et al.}~\cite{tai2024defect} introduced StableSDG, which adapts Stable Diffusion~\cite{rombach2022high} for generating defect images to aid in training recognition models.
Although these diffusion-based methods produce high-quality images, they lack control over pixel-level labels. To enhance control, some works~\cite{chen2024training,xie2023boxdiff,li2023gligen,Wang_2024_CVPR,endo2024masked} focus on conditioning diffusion models with grounding inputs like bounding boxes, keypoints, or edge maps. However, these methods mainly generate natural content, such as animals and scenes, matching Stable Diffusion's training data, which differs significantly from defect image distribution.

Thus, it is crucial to design a pixel-level controlled generation method for saliency-based defect detection. We introduce DefFiller, a new approach that builds on the pre-trained GLIGEN model~\cite{li2023gligen} to generate mask-conditioned defects on steel surfaces. We also present a thorough evaluation framework to measure the quality of the generated samples and their impact on detection accuracy.

The main contributions of this paper are as follows:
\begin{itemize}
    \item[1)]To enhance saliency-based defect detection models, we introduce DefFiller, a defect generation method that combines the broad knowledge of a pre-trained diffusion model with mask conditions through additional trainable layers. To our knowledge, this is the first approach to blend diffusion priors to mask-conditioned defect generation.
    
    \item[2)]We develop a thorough evaluation framework for mask-conditioned generative models. This framework evaluates the quality of the generated samples and integrates these mask-image pairs into training advanced saliency-based defect detection models, comparing their performance before and after the data expansion.
        
    \item[3)]Compared to existing methods, our approach achieves high quality and strong controllability in mask-conditioned defect generation. Experimental results also demonstrate our method's effectiveness and superiority in dataset expansion.
    
\end{itemize}

\section{Related work}
\label{sec2}

In this section, we review existing work on defect image generation, including both GAN-based and diffusion-based methods. We also discuss advancements in saliency-based defect detection, which we use to evaluate the effectiveness of the generated defect images in this paper.

\subsection{GAN-based defect generation}

Generative Adversarial Networks (GANs)~\cite{goodfellow2014generative} are foundational generative models known for their powerful image generation capabilities. Variants of GANs, such as CycleGAN~\cite{zhu2017unpaired} and StyleGAN~\cite{sauer2022stylegan}, have been widely applied to various image generation tasks, including photo cartoonization~\cite{zhao2024gan} and multi-stitch embroidery generation~\cite{hu2024msembgan}.
Several defect generation methods build on these GAN variations. For example, Zhang~\textit{et al.}~\cite{zhang2021defect} propose a compositional layer-based approach to generate and remove defects in surface images. Similarly, Zhao~\textit{et al.}~\cite{zhao2023defect} integrate transformer and U-Net models to capture both global and local features, enabling the transformation of defect-free images into defective ones.

To control the regions of generated defects, Li~\textit{et al.}~\cite{li2023dls} use two Encoder-Decoder models to extract defect features and locations from both defect and defect-free images, then combine these features to synthesize images with specific defects.
Duan~\textit{et al.}~\cite{Duan2023DFMGAN} propose a two-stage approach: first, they pre-train the model on defect-free images, then adapt it to generate realistic defect masks and images.
Ran~\textit{et al.}~\cite{ran2024sketch} extract defect edges and background texture from the original image and use these as inputs to the network, improving the quality of the generated images.
However, training these generative models often requires starting from scratch and depends on having a sufficient number of defect-free images. When image samples are limited, it can result in undesirable patterns in the generated outputs.

\subsection{Diffusion-based defect generation}


With efficient generation from latent space, the Stable Diffusion model~\cite{rombach2022high} has been used in various tasks, including image generation~\cite{balaji2022ediffi,ramesh2022hierarchical}, 3D generation~\cite{ma2024diffspeaker,ma2025scaledreamer}, and image super-resolution~\cite{wu2024one}. Because of its high-fidelity results, some studies have begun to explore using diffusion-based models to generate defect data.
Yang~\textit{et al.}~\cite{yang2024novel} train a DDPM~\cite{ho2020denoising} to create samples for fault diagnosis.
Xiao~\textit{et al.}~\cite{xiao2024parameter} incorporate a parameter-sharing attention mechanism into the diffusion process to generate mechanical fault samples.
Tai~\textit{et al.}~\cite{tai2024defect} develop a pipeline to adapt text-to-image generative models for producing defect image samples. While these methods can generate high-quality defect samples, the defect regions often appear random and uncontrolled.

Recently, several studies~\cite{xie2023boxdiff,endo2024masked,chen2024training,li2023gligen} have explored adapting Stable Diffusion to handle layout conditions, allowing for more control over the generation regions. 
For example, Zhang~\textit{et al.}~\cite{zhang2024ag} introduce an aquascape generation method that leverages ControlNet~\cite{zhang2023adding} to regulate the overall structure of the generated aquascape images.
However, these methods primarily focus on generating natural content, which differs significantly from the distribution of defect images.
To tackle this challenge, Li~\textit{et al.}~\cite{li2024novel} develop an inference strategy based on the Blended Latent Diffusion Model~\cite{avrahami2023blended} for industrial anomaly detection. However, this approach's reliance on defect-free images poses challenges for practical applications.
Therefore, we propose DefFiller, a mask-conditioned defect generation method with diffusion prior to achieve pixel-level control.

\subsection{Saliency-based defect detection}

Saliency detection~\cite{achanta2009frequency} can capture the subset of vital visual information of image for further processing and filter out plenty of redundant background interferences. Coupled with the rapid development of deep learning technologies, saliency-based defect detection models~\cite{sun2024adversarial,ding2022cross,shen2024minet,ma2023skin} present promising detection results. 
Some of these efforts~\cite{han2022two,zhou2021dense,zhou2021edge,dong2019pga} focus on adopting the multi-scale strategy to provide rich contextual information for defects regions and overcome the challenges of defect detection. Zhou~\textit{et al.}~\cite{zhou2021dense} propose a dense attention-guided cascaded network (DACNet) by deploying multi-resolution convolutional branches. Zhou~\textit{et al.}~\cite{zhou2021edge} further propose an edge-aware multilevel interactive network, which relies on the interactive feature integration and the edge-guided saliency fusion. Besides, Han~\textit{et al.}~\cite{han2022two} design a two-stage edge reuse network, which executes prediction and refinement successively.

However, these methods often face significant computational overhead and slow processing speeds, which strain the limited storage and processing power of industrial devices. As a result, many researchers~\cite{shen2024minet,wan2023sminet} are focusing on developing lightweight, saliency-based networks for defect detection.
In this paper, we use CSEPNet~\cite{ding2022cross}, TSERNet~\cite{han2022two} and MINet~\cite{shen2024minet} for our experiments. We generate virtual mask-image pairs with a generative model to expand the dataset and train the detection model on both the original and expanded datasets. We then evaluate the effectiveness of the generative model by comparing the detection performance.

\section{Method}
\label{sec3}

In this section, we first present DefFiller for mask-conditioned defect generation, which leverages the GLIGEN model~\cite{li2023gligen} as the diffusion prior. We also introduce our evaluation framework to assess the quality of the generated samples and their effect on enhancing detection performance.

\begin{figure}[t]
    \centering
    \includegraphics[width=1.0\linewidth]{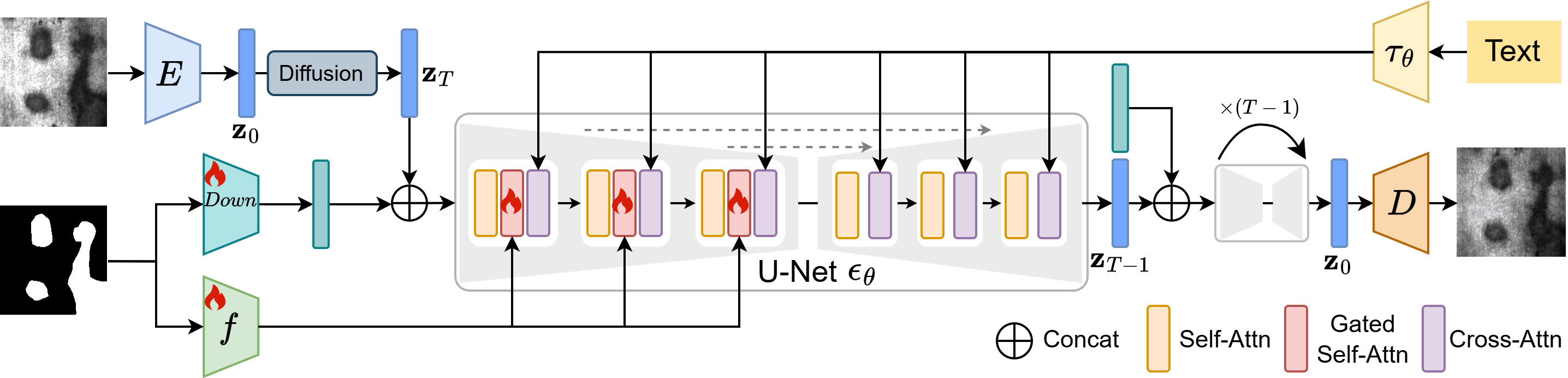}
    \caption{
    Overview of DefFiller. During training, only the parameters in the gated self-attention layers, the mask encoder and the downsmpling network are optimized. At inference, a random noise tensor $\mathbf{z}_T$ is sampled from a standard Gaussian distribution. With guidance from the text prompt, DefFiller generates defect samples that match the mask conditions.
    }  
    \label{fig_method}
\end{figure}

\subsection{DefFiller}

The diffusion model~\cite{ho2020denoising} is a type of generative model that learns data distribution by gradually adding noise and then recovering the original data. For layout-to-image generation, GLIGEN~\cite{li2023gligen} is widely used, combining an autoencoder with a layout-conditioned latent diffusion model. Our method builds on this framework by adding an encoder for mask conditions. An overview of DefFiller is shown in Fig.~\ref{fig_method}. Next, we will delve into the key components of DefFiller.

\subsubsection{Auto-encoder} 
To perform diffusion and denoising in the low-dimensional latent space, an autoencoder transforms images into latent codes and vice versa. The encoder \(E(\cdot)\) maps images \(\mathbf{x} \in \mathbb{R}^D\) into latent codes \(\mathbf{z}=E(\mathbf{x})\), where \(\mathbf{z} \in \mathbb{R}^K\) and \(K \ll D\). The decoder \(D(\cdot)\) then converts these latent codes back into images. With sufficient training, it holds that \(D(E(\mathbf{x})) \approx \mathbf{x}\).

\subsubsection{Mask encoder} 
\begin{figure}[t]
    \centering
    \includegraphics[width=1.0\linewidth]{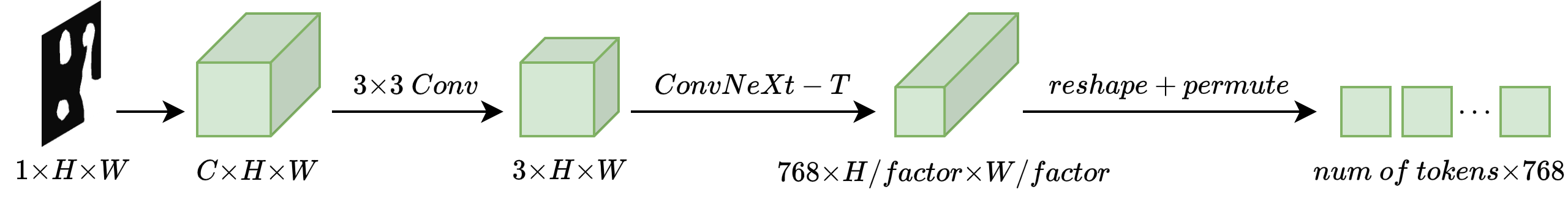}
    \caption{
    The architecture of the mask encoder.
    }  
    \label{fig_mask_encoder}
\end{figure}

The mask $\mathbf{e}$ acts as a semantic map, with layout information embedded in each spatial coordinate. We design a network $f(\cdot)$ to convert the mask into layout tokens $\boldsymbol{h}^\textit{e}$, its architecture is shown in Fig.~\ref{fig_mask_encoder}. First, we expand the mask channels to match the number of semantic classes. Then, using a $3\times3~Conv$ layer followed by the pre-trained $ConvNeXt-T$~\cite{liu2022convnet}, we extract mask features. These features are reshaped and permuted to produce layout tokens. We adopt GLIGEN's settings: $C=152$, $H=W=256$, $factor=32$, and $num~of~tokens=64$.

\subsubsection{Layout-conditioned latent diffusion model} 

The latent diffusion model typically uses a U-Net structure, denoted as $\epsilon_\theta(\cdot)$, which consists of several encoders and decoders. Each encoder includes a residual block, a self-attention layer, and a cross-attention layer in sequence. For a given timestep $t$, a text prompt $\boldsymbol{c}$, and a mask $\mathbf{e}$ as conditions, the layout-conditioned latent diffusion model can be represented as $\hat{\epsilon} =  \epsilon_{\theta}(\mathbf{z}_t, (\boldsymbol{h}^\textit{c},\boldsymbol{h}^\textit{e}), t)$. Here, $\boldsymbol{h}^\textit{c}$ represents the prompt tokens encoded by a fixed text encoder $\tau_{\theta}(\cdot)$~\cite{radford2021learning}, computed as $\boldsymbol{h}^\textit{c} = \tau_{\theta}(\boldsymbol{c})$, and $\boldsymbol{h}^\textit{e}$ denotes the layout tokens extracted from the mask conditions, computed as $\boldsymbol{h}^\textit{e} = f(\mathbf{e})$.

To incorporate the new layout information, GLIGEN adds a gated self-attention layer between the self-attention and cross-attention layers within the encoder by introducing new parameters. For an input image feature $\boldsymbol{v}$, the processing within each attention block is as follows:
\begin{align}
\boldsymbol{v} &= \boldsymbol{v} + \text{SelfAttn}(\boldsymbol{v}), \\
\boldsymbol{v} &= \boldsymbol{v} + \tanh (\gamma) \cdot \text{TS}\left(\text{Gated-SelfAttn}\left(\left[\boldsymbol{v}, \boldsymbol{h}^\textit{e}\right]\right)\right), \\
\boldsymbol{v} &= \boldsymbol{v} + \text{CrossAttn}(\boldsymbol{v}, \boldsymbol{h}^\textit{c}),
\end{align}
where $\text{TS}(\cdot)$ denotes a token selection operation that retains only the image feature after applying $\text{Gated-SelfAttn}(\cdot)$, and $\gamma$ is a learnable scalar initially set to 0.

\subsubsection{Training objective} 
Given a dataset of mask-image pairs, the pre-trained auto-encoder, text encoder, and layout-conditioned latent diffusion model, the training objective for DefFiller is defined by the following loss function:
\begin{equation}
\begin{aligned}
\label{layout_diffusion_training_objective}
    \mathcal{L}_{\text{DefFiller}} &= \mathbb{E}_{t, \epsilon, \mathbf{x}, \mathbf{z} = E(\mathbf{x})}\left[\left\|\epsilon_{\phi}(\mathbf{z}_t, (\tau_{\theta}(\boldsymbol{c}), f_{\phi}(\mathbf{e})), t) - \epsilon\right\|_2^2\right].
\end{aligned}
\end{equation}
This function calculates the Mean Squared Error between the predicted noise and the actual noise, averaged over the text prompt $\boldsymbol{c}$, mask-image pairs $(\mathbf{x},\mathbf{e}) \sim S$, noise $\epsilon \sim \mathcal{N}(\mathbf{0}, \boldsymbol{I})$, and timestep $t \sim \{1, \cdots, T\}$.
The model is trained by adjusting all parameters denoted by $\phi$, which includes parameters in the mask encoder $f_{\phi}(\cdot)$ and the gated self-attention layers. Additionally, to speed up training, $\mathbf{e}$ is fed into the first convolutional layer of the U-Net. Specifically, the input to the U-Net is $\text{CONCAT}({Down}_{\phi}(\mathbf{e}), \mathbf{z}_t)$, where ${Down}_{\phi}(\cdot)$ is a downsampling network that reduces $\mathbf{e}$ to match the spatial resolution of $\mathbf{z}_t$. In this paper, the first convolutional layer of the U-Net is also trainable.

Additionally, the model employs classifier-free guidance~\cite{ho2022classifier} to balance fidelity and diversity:
\begin{equation}
    \label{classifier_free_guidance}
    \hat{\epsilon} = \epsilon_{\phi}(\mathbf{z}_{in}, f_{\phi}(\mathbf{e}), t) + \omega_{cfg} \left[\epsilon_{\phi}(\mathbf{z}_{in}, (\tau_{\theta}(\boldsymbol{c}), f_{\phi}(\mathbf{e})), t) - \epsilon_{\phi}(\mathbf{z}_{in}, f_{\phi}(\mathbf{e}), t)\right],
\end{equation}
where $\mathbf{z}_{in}=\text{CONCAT}({Down}_{\phi}(\mathbf{e}),\mathbf{z}_t)$, $\epsilon_{\phi}(\mathbf{z}_{in}, f_{\phi}(\mathbf{e}),t)$ represents the noise prediction without text prompt guidance, and $\omega_{cfg}$ is a scalar that controls the impact of the text condition on the generation process.
During inference, a random noise tensor is sampled and iteratively denoised to generate new latent codes $\mathbf{z}_0$, which are then decoded into an image via $\mathbf{x} = D(\mathbf{z}_0)$.

\subsection{Evaluation framework}

To thoroughly assess the generative models for mask-conditioned steel surface defect generation, we design a evaluation framework that emphasizes both the quality of the generated samples and their effectiveness in enhancing detection performance.

\subsubsection{Generation quality}

We use the Fréchet Inception Distance (FID) metric \cite{heusel2017gans} to assess the quality of generated defect images. FID measures the similarity between real and generated image distributions with the formula:
\begin{equation}
\mathrm{FID}={\left\|\mu_r-\mu_g\right\|}^2+\operatorname{Tr}\left(C_r+C_g-2\left(C_r C_g\right)\right)^{1 / 2},
\label{FID}
\end{equation}
where $\mu_r$ and $\mu_g$ are the mean feature vectors for real and generated images, respectively, and $C_r$ and $C_g$ are their covariance matrices.
To enhance the quality of the generated defect images, we iteratively adjust the guidance scale $\omega_{cfg}$ for each defect category to achieve lower FID scores, as detailed in Table~\ref{ablatin_study_gs}.

\subsubsection{Detection performance}

After evaluating quality, we select the mask-image pairs that best match the optimal distribution to expand the dataset. We then train saliency-based detection models, CSEPNet~\cite{ding2022cross}, TSERNet~\cite{han2022two} and MINet~\cite{shen2024minet}, on this enhanced defect dataset.
To assess the performance of these detection models, we employ four metrics, including S-measure~\cite{fan2017structure}, mean absolute error, E-measure~\cite{fan2018enhanced} and F-measure~\cite{achanta2009frequency}. 
\begin{itemize}
    \item[1)] S-measure ($S_\alpha$, with $\alpha=0.5$) assesses structural similarity at both object and region levels.
    \item[2)] Mean Absolute Error (MAE, $\mathcal{M}$) measures the difference between predictions and ground truth at the pixel level.
    \item[3)] E-measure ($E_{\xi}$) evaluates image-level statistics and local pixel matching, with the maximum form $E_{\xi}^{\max}$ used in our experiment.
    \item[4)] F-measure ($F_\beta$) focuses on boundary quality and edge details, with the maximum form $F_\beta^{\max}$ used in our experiment.
    
\end{itemize}

\section{Experiment}
\label{sec4}

\subsection{Experimental setting}

\begin{figure}[t]
    \centering
    \includegraphics[width=0.6\linewidth]{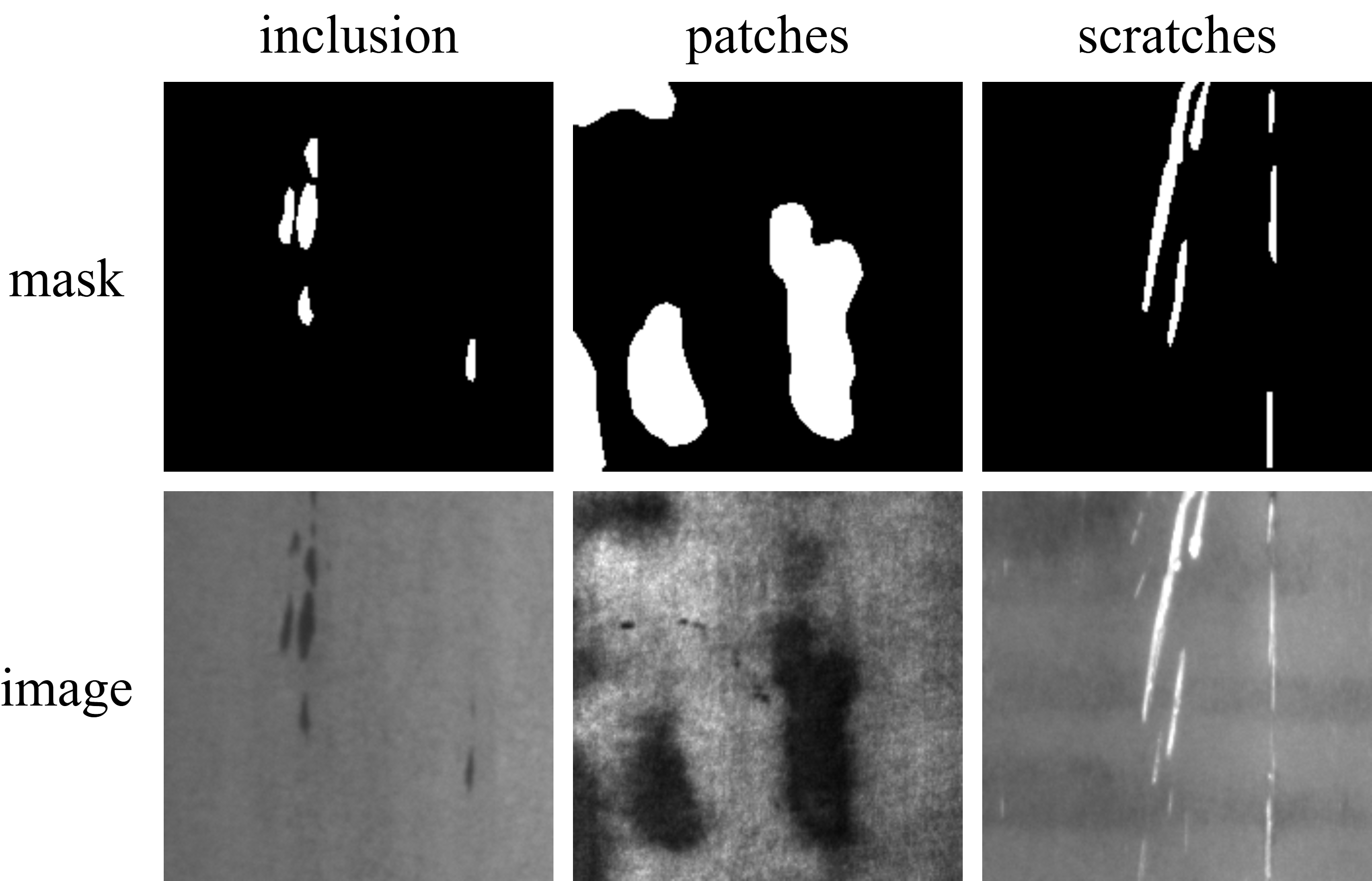}
    \caption{The illustration of mask-image pairs in SD-Saliency-900 dataset.}  
    \label{fig:dataset}
\end{figure}

\subsubsection{Dataset} To thoroughly evaluate DefFiller, we perform both quantitative and qualitative tests using the SD-Saliency-900 dataset~\cite{song2020edrnet}, which is a challenging dataset for strip steel surface defect detection. This dataset includes three defect types—inclusions, patches, and scratches—with each type having 300 mask-image pairs at a resolution of 200 × 200, as partially shown in Fig.~\ref{fig:dataset}.

\subsubsection{Implementation details} Our method starts by loading weights from the GLIGEN model~\cite{li2023gligen} and then fine-tunes the parameters in the gated self-attention layers, the mask encoder and the downsmpling network using the Adam optimizer~\cite{kingma2014adam} with mask-image pairs from the SD-Saliency-900 dataset. This optimization process runs for 30,000 iterations, including a 10,000-iteration warm-up period. The learning rate is initially set to 5e-5, and the batch size is 2. 
For the comparison methods, we adhere to their official implementations and train the model to ensure the best possible performance.
In the experiments described in Sections~\ref{ablation_study} and~\ref{data_substitution}, we use the ground truth from the dataset as the mask condition, with each generative model producing 300 defect images per category. In Section~\ref{data_expansion}, we generate 900 new masks to guide the process, enhancing the diversity of defect samples.
All experiments are conducted with 1 NVIDIA A6000 GPU.

\subsection{Ablation study}
\label{ablation_study}

\subsubsection{Training strategy} 
\begin{figure}[t]
    \centering
    \includegraphics[width=0.9\linewidth]{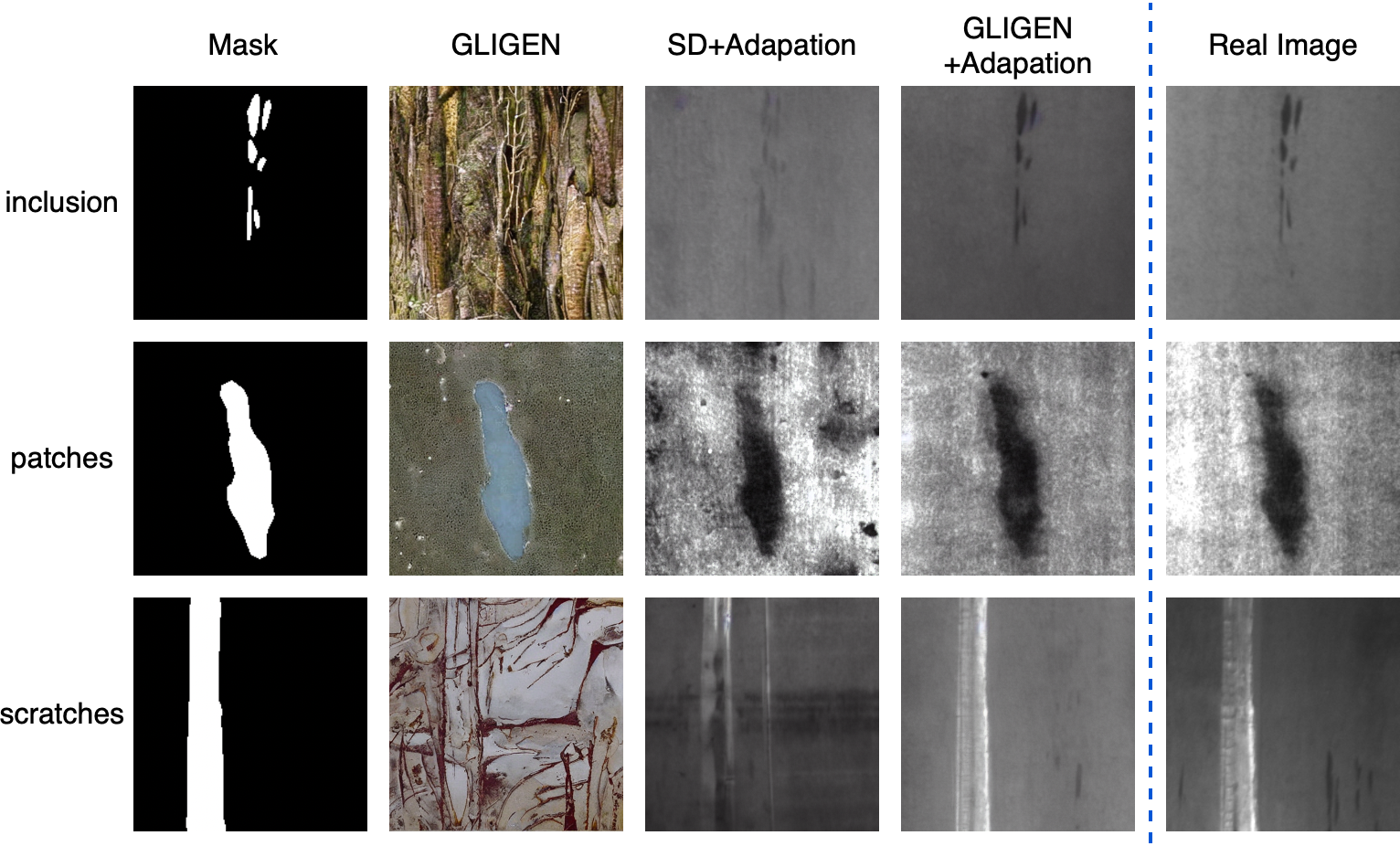}
    \caption{Visualization of generated images with different training strategies.}  
    \label{deffiller_abla_1}
\end{figure}

To preserve the diffusion model's general knowledge, we first load pre-trained weights and then adapt it using mask-image pairs from the steel surface defect dataset. Table~\ref{ablatin_study_strategy} and Fig.~\ref{deffiller_abla_1} show the quantitative and qualitative evaluation of defect images generated by the model under various training strategies. The results indicate that: 1) adapting to the defect dataset enables the model to understand defect-specific concepts, differentiating them from natural content; and 2) the GLIGEN model~\cite{gligen-sem}, pre-trained on the ADE20K dataset~\cite{zhou2017scene}, provides initialization parameters for mask extraction, leading to a significant improvement in the quality of generated images compared to SD v1.4~\cite{stable-diffusion-v1-4}. Thus, DefFiller adopts the optimal strategy (the third line) for better generation performance.

\begin{table}[t]
\caption{FID scores of generated images with different training strategies.}
\label{ablatin_study_strategy}
\renewcommand\arraystretch{1.2}
\begin{tabular*}{\textwidth}{@{\extracolsep\fill}lccccccc}
\toprule%
\multicolumn{2}{@{}c@{}}{Load pre-trained weights} & \multirow{2}*{Adapation} & \multicolumn{3}{@{}c@{}}{Defect Category}& \multirow{2}*{AVG} \\\cmidrule{1-2} \cmidrule{4-6}
SD v1.4~\cite{stable-diffusion-v1-4}&GLIGEN~\cite{gligen-sem}&&inclusion&patches&scratches&\\
\midrule
&\checkmark&&394.96&365.68&313.57&358.07\\
\checkmark&&\checkmark&65.63&83.87&74.05&74.52\\
&\checkmark&\checkmark&47.03&63.79&52.37&\textbf{54.40}\\
\botrule
\end{tabular*}
\end{table}

\subsubsection{Guidance scale} 
\begin{figure}[t]
    \centering
    \includegraphics[width=0.98\linewidth]{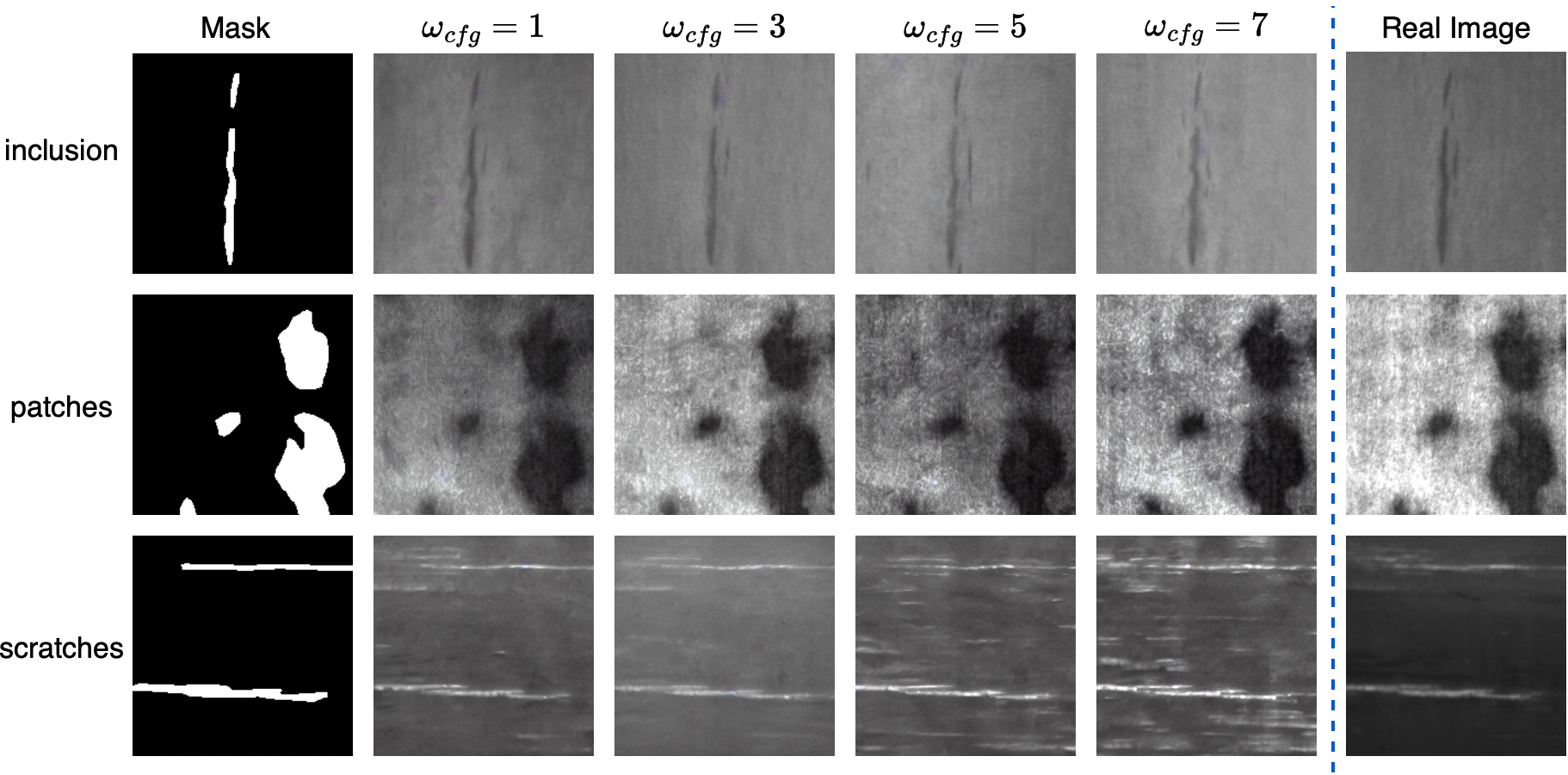}
    \caption{Visualization of generated images with different guidance scale $\omega_{cfg}$.}  
    \label{deffiller_abla_2}
\end{figure}

We emphasize the significance of the guidance scale $\omega_{cfg}$ in classifier-free guidance \cite{ho2022classifier}. As illustrated in Table \ref{ablatin_study_gs}, there is approximately a 14\% improvement in average FID metrics between the best and worst performances, indicating that the optimal guidance scale greatly enhances the quality of generated images. Along with the visualizations in Fig.~\ref{deffiller_abla_2}, we set $\omega_{cfg}$ to 3 in this paper.

\begin{table}[t]
\caption{FID scores of generated images with different guidance scale $\omega_{cfg}$.}
\label{ablatin_study_gs}
\renewcommand\arraystretch{1.2}
\begin{tabular*}{\textwidth}{@{\extracolsep\fill}lccccc}
\toprule%
\multirow{2}*{Guidance Scale $\omega_{cfg}$}& \multicolumn{3}{@{}c@{}}{Defect Category} & \multirow{2}*{AVG} \\\cmidrule{2-4}%
&inclusion&patches&scratches&  \\
\midrule
    1&46.16&63.80&55.01&54.99\\
    3&47.03&63.79&52.37&\textbf{54.40}\\
    5&46.89&63.84&55.82&55.52\\
    7&53.25&70.98&66.29&63.51\\
\botrule
\end{tabular*}
\end{table}

\subsection{Data substitution}
\label{data_substitution}

To assess the fidelity of the generative models, we first use the initial masks from the SD-Saliency-900 dataset as conditions. We then perform quantitative and qualitative experiments on the generated defect images, evaluating both the similarity between the generated samples and real images, as well as the performance changes in the detection model before and after substituting the training data.
It is important to note that DFMGAN~\cite{Duan2023DFMGAN} generates defect images and corresponding masks simultaneously, without being controlled by a given mask condition. Therefore, we will not include it in this section for comparison.

\begin{table}[t]
\caption{FID scores for defect images generated from the masks in SD-Saliency-900.}
\label{substitution_fid}
\renewcommand\arraystretch{1.2}
\begin{tabular*}{\textwidth}{@{\extracolsep\fill}lccccc}
\toprule%
\multirow{2}*{Method}& \multicolumn{3}{@{}c@{}}{Defect Category} & \multirow{2}*{AVG} \\\cmidrule{2-4}%
&inclusion&patches&scratches&  \\
\midrule
    DFMGAN~\cite{Duan2023DFMGAN}&-&-&-&-\\
    AdaBLDM~\cite{li2024novel}&99.75&144.66&74.25&106.22\\
    DefFiller (Ours)&\textbf{47.03}&\textbf{63.79}&\textbf{52.37}&\textbf{54.40}\\
\botrule
\end{tabular*}
\end{table}

\subsubsection{Generation quality} 
\label{substitution_quality}

\begin{figure}[t]
    \centering
    \includegraphics[width=0.85\linewidth]{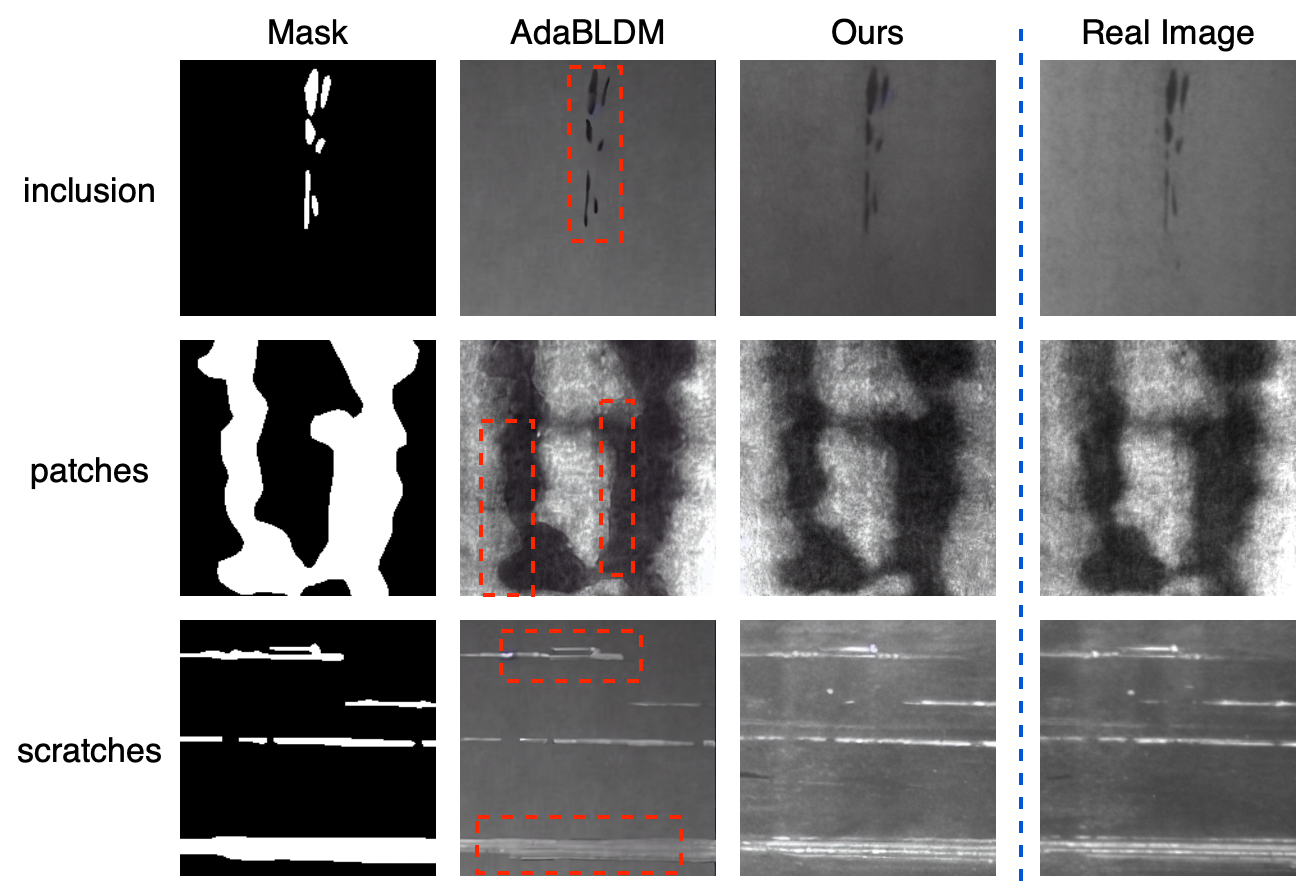}
    \caption{Qualitative comparison of generated defect images, of which the mask conditions are from the SD-Saliency-900 dataset. Artifacts are highlighted in red boxes, with details in Section~\ref{substitution_quality}.
    }  
    \label{substitution_visulization}
\end{figure}

Table~\ref{substitution_fid} presents the image generation performance, as measured by FID, in three categories of the dataset. Qualitative comparisons are shown in Fig.~\ref{substitution_visulization}, where regions with abnormal textures in the generated images are highlighted in red boxes. It is evident that AdaBLDM produces images with a noticeable difference between defect features and background, resulting in unnatural transitions. In contrast, DefFiller generates high-quality defect samples, achieving the lowest FID scores in each category. Its visualization results further demonstrate that the generated samples closely resemble real images and align well with the given mask conditions.

\begin{table}[t]
\caption{Parameters settings of detection models.}
\label{detection_setting}
\renewcommand\arraystretch{1.2}
\begin{tabular*}{\textwidth}{@{\extracolsep\fill}lcccc}
\toprule%
&CSEPNet~\cite{ding2022cross}&TSERNet~\cite{han2022two}&MINet~\cite{shen2024minet}\\
\midrule
    Learning rate&1e-3&1e-3&4e-3\\
    Optimizer&SGD&Adam&Adam\\
    Weight decay&5e-4&0&5e-4\\
    Momentum&0.9&0.9&0.9\\
    Batch size&4&4&32\\
\botrule
\end{tabular*}
\end{table}

\begin{figure}[t]
    \centering
    \includegraphics[width=0.68\linewidth]{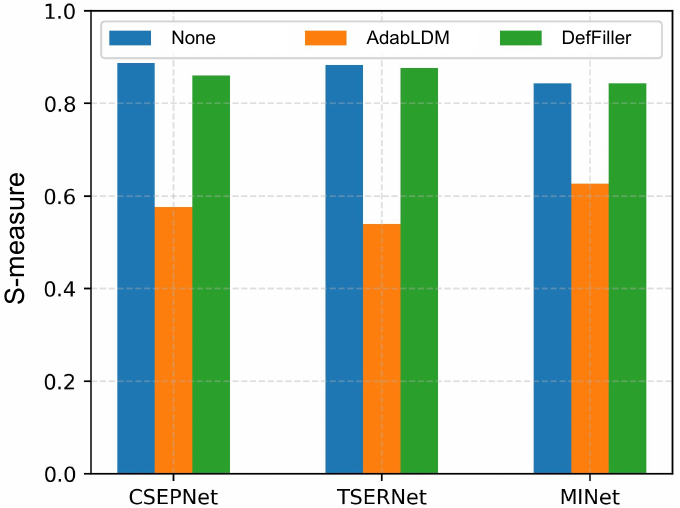}
    \caption{Comparison on S-measure before and after data substitution.
    }  
    \label{substitution_plot}
\end{figure}

\begin{table}[t]
\caption{Performance of defect detection models before and after data substitution.}
\label{comparison_substitution}
\renewcommand\arraystretch{1.2}
\begin{tabular*}{\textwidth}{@{\extracolsep\fill}lcccccc}
\toprule%
Network&Generation method&$S_\alpha \uparrow$&$\mathcal{M} \downarrow$&$E_{\xi}^{\max } \uparrow$&$F_\beta^{\max } \uparrow$\\
\midrule
    \multirow{3}*{\makecell*[c]{CSEPNet~\cite{ding2022cross}}}&None&0.887&0.023&0.968&0.894\\
    &AdaBLDM~\cite{li2024novel}&0.576&0.115&0.540&0.343\\
    &DefFiller (Ours)&0.860&0.027&0.954&0.864\\
\midrule
    \multirow{3}*{\makecell*[c]{TSERNet~\cite{han2022two}}}&None&0.883&0.024&0.965&0.892\\
    &AdaBLDM~\cite{li2024novel}&0.540&0.121&0.669&0.367\\
    &DefFiller (Ours)&0.877&0.026&0.961&0.881\\
\midrule
    \multirow{3}*{\makecell*[c]{MINet~\cite{shen2024minet}}}&None&0.843&0.036&0.934&0.824\\
    &AdaBLDM~\cite{li2024novel}&0.627&0.163&0.686&0.546\\
    &DefFiller (Ours)&0.843&0.034&0.945&0.835\\
\botrule
\end{tabular*}
\end{table}

\subsubsection{Detection performance}
\label{substitution_detection}

We also use the generated defect images for data substitution and assess how the detection model's performance changes before and after this substitution. The SD-Saliency-900 dataset is split into a 9:1 ratio of training and testing sets. The training parameters of detection models are detailed in Table~\ref{detection_setting}. We train CSEPNet~\cite{ding2022cross}, TSERNet~\cite{han2022two}, and MINet~\cite{shen2024minet} until convergence, following official guidelines. The comparison of S-measure performance is shown in Fig.~\ref{substitution_plot}.
The blue bars represent the original performance on the testing set. The orange and green bars indicate the model performance after replacing the initial training set with defect images generated by AdaBLDM and DefFiller, respectively. We observe that replacing the training set with AdaBLDM-generated images significantly degrades the model's performance, while using images from DefFiller maintains performance similar to the original. This demonstrates that our generated images closely match real ones and are highly consistent with the original mask annotations. A comparison of other metrics is provided in Table~\ref{comparison_substitution}, further emphasizing the high fidelity of defect samples produced by DefFiller.

\subsection{Data expansion}
\label{data_expansion}

To evaluate how well the model handles new mask conditions and how generated samples enhance detection performance, we create new masks to generate defect images. Then the generated 300 mask-image pairs per category are used to expand the training set for the detection models. It is worth noting that DFMGAN generates both masks and defect samples simultaneously, while AdaBLDM and DefFiller use masks created by DDPM~\cite{ho2020denoising} as conditions for generating defect images.

\begin{table}[t]
\caption{FID scores for masks generated by DDPM.}
\label{mask_fid}
\renewcommand\arraystretch{1.2}
\begin{tabular*}{\textwidth}{@{\extracolsep\fill}lccccc}
\toprule%
\multirow{2}*{Method}& \multicolumn{3}{@{}c@{}}{Defect Category} & \multirow{2}*{AVG} \\\cmidrule{2-4}%
&inclusion&patches&scratches&  \\
\midrule
    DDPM~\cite{ho2020denoising}&96.61&99.72&94.46&96.93\\
\botrule
\end{tabular*}
\end{table}

\begin{figure}[t]
    \centering
    \includegraphics[width=1.0\linewidth]{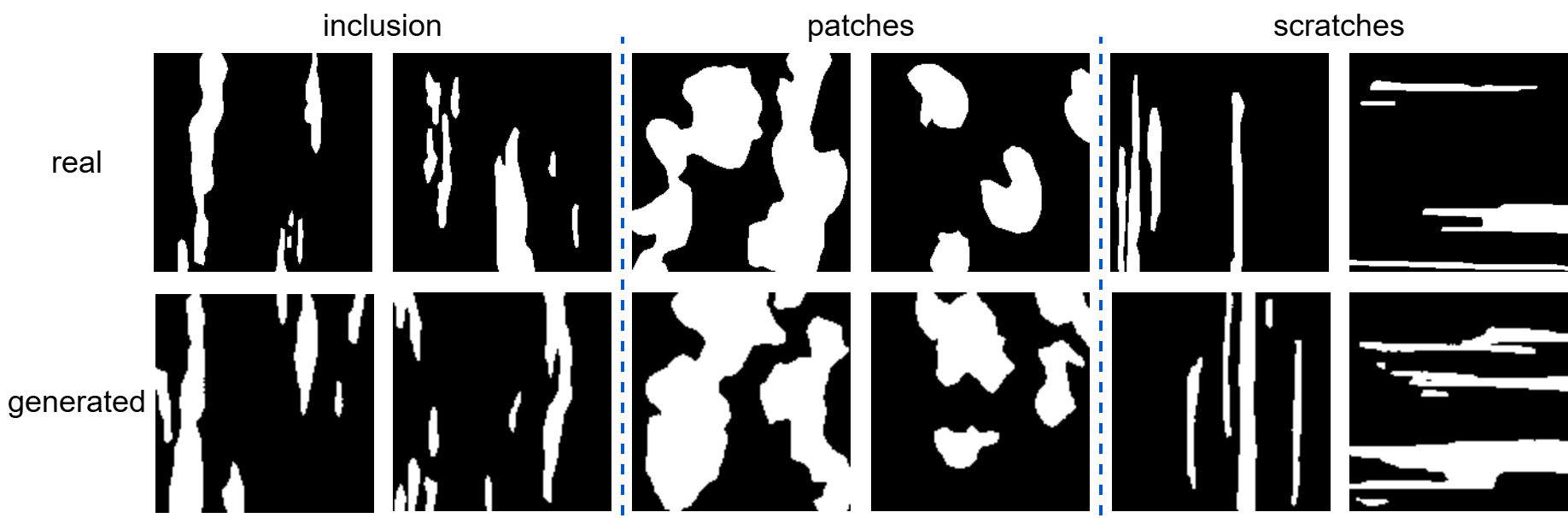}
    \caption{Visualization results of masks generated by DDPM.
    }  
    \label{mask_ddpm}
\end{figure}

\subsubsection{Mask producer}
\label{mask_producer}

We train a DDPM from scratch for 200 epochs using the ground truth from the SD-Saliency-900 dataset. The training is done with a batch size of 32 and a learning rate of 1e-4, at a resolution of 64 $\times$ 64. After binarizing and resizing the generated images to 256 × 256, we create 300 new masks per category for data expansion. To evaluate the mask quality, we calculate FID scores by comparing the generated masks to real ones for each category, as shown in Table~\ref{mask_fid}. Visualization results are in Fig.~\ref{mask_ddpm}. The high-quality generated masks closely resemble the real ones. We then use these masks as conditions to guide AdaBLDM and DefFiller in producing defect samples.

\subsubsection{Generation quality}
\label{expansion_quality}

\begin{figure}[t]
    \centering
    \includegraphics[width=1.0\linewidth]{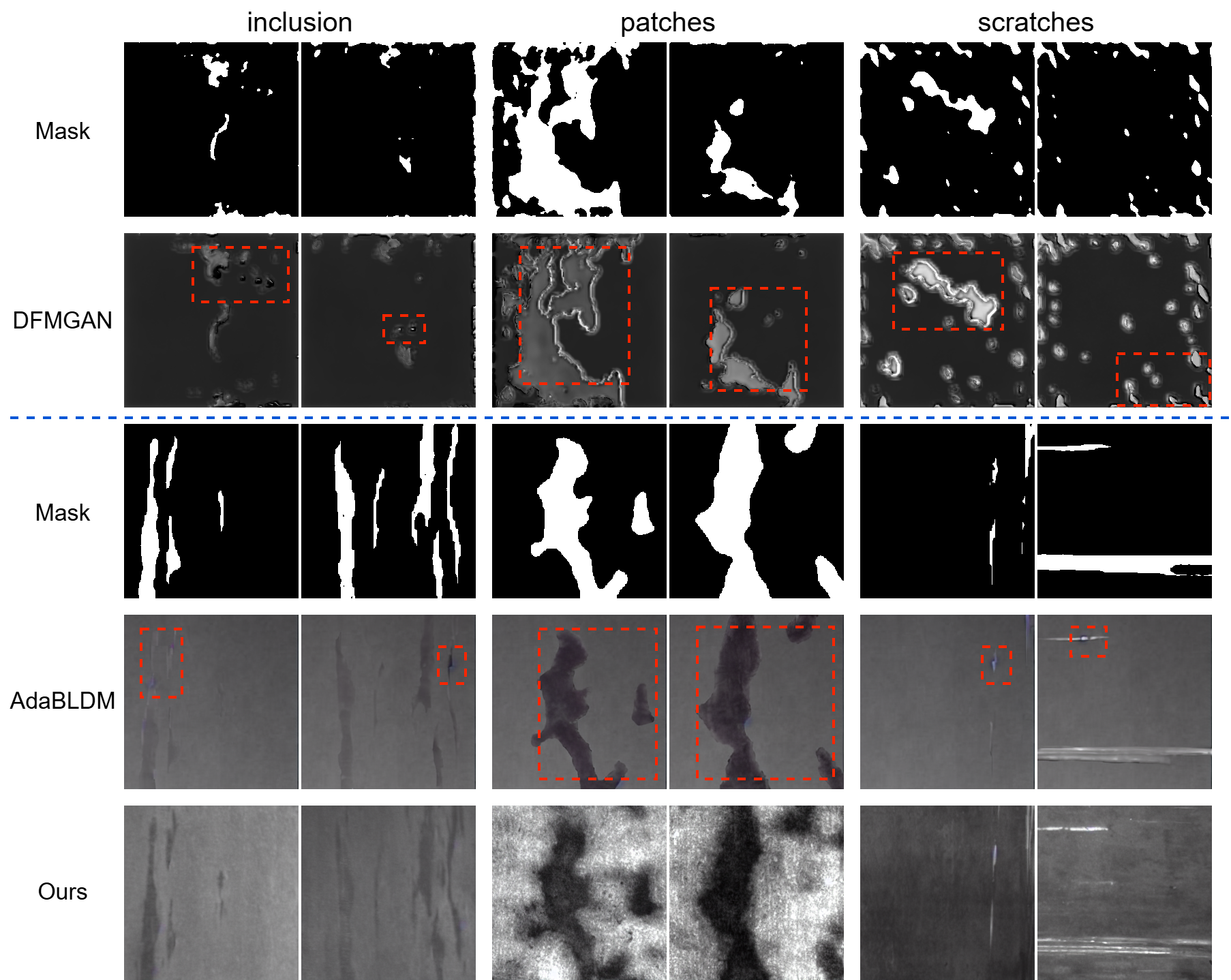}
    \caption{Qualitative comparison of generated defect images. DFMGAN generates masks and images simultaneously, while AdaBLDM and DefFiller adopt the masks produced by DDPM as conditions. Artifacts are highlighted in red boxes, with details in Section~\ref{expansion_quality}.
    }  
    \label{expansion_visulization}
\end{figure}

We evaluate the model's generation quality through both quantitative and qualitative experiments. Table \ref{expansion_fid} shows the FID scores for defect images generated from the new masks, and Fig. \ref{expansion_visulization} presents visualization results.
The DFMGAN-generated masks and images are consistent, but the defect images exhibit patches of uneven brightness, differing significantly from real images. AdaBLDM, which depends on defect-free images for backgrounds, struggles when such images are scarce, resulting in unrealistic and unnatural defect features pasted onto the background.
In contrast, our method achieves the lowest FID scores while adhering to mask conditions.

\begin{table}[t]
\caption{FID scores for defect images generated from the newly produced masks.}
\label{expansion_fid}
\renewcommand\arraystretch{1.2}
\begin{tabular*}{\textwidth}{@{\extracolsep\fill}lccccc}
\toprule%
\multirow{2}*{Method}& \multicolumn{3}{@{}c@{}}{Defect Category} & \multirow{2}*{AVG} \\\cmidrule{2-4}%
&inclusion&patches&scratches&  \\
\midrule
    DFMGAN~\cite{Duan2023DFMGAN}&376.41&362.60&414.61&384.54\\
    AdaBLDM~\cite{li2024novel}&282.47&363.57&245.60&297.21\\
    DefFiller (Ours)&\textbf{74.19}&\textbf{102.23}&\textbf{86.62}&\textbf{87.68}\\
\botrule
\end{tabular*}
\end{table}

\begin{figure}[t]
    \centering
    \includegraphics[width=0.7\linewidth]{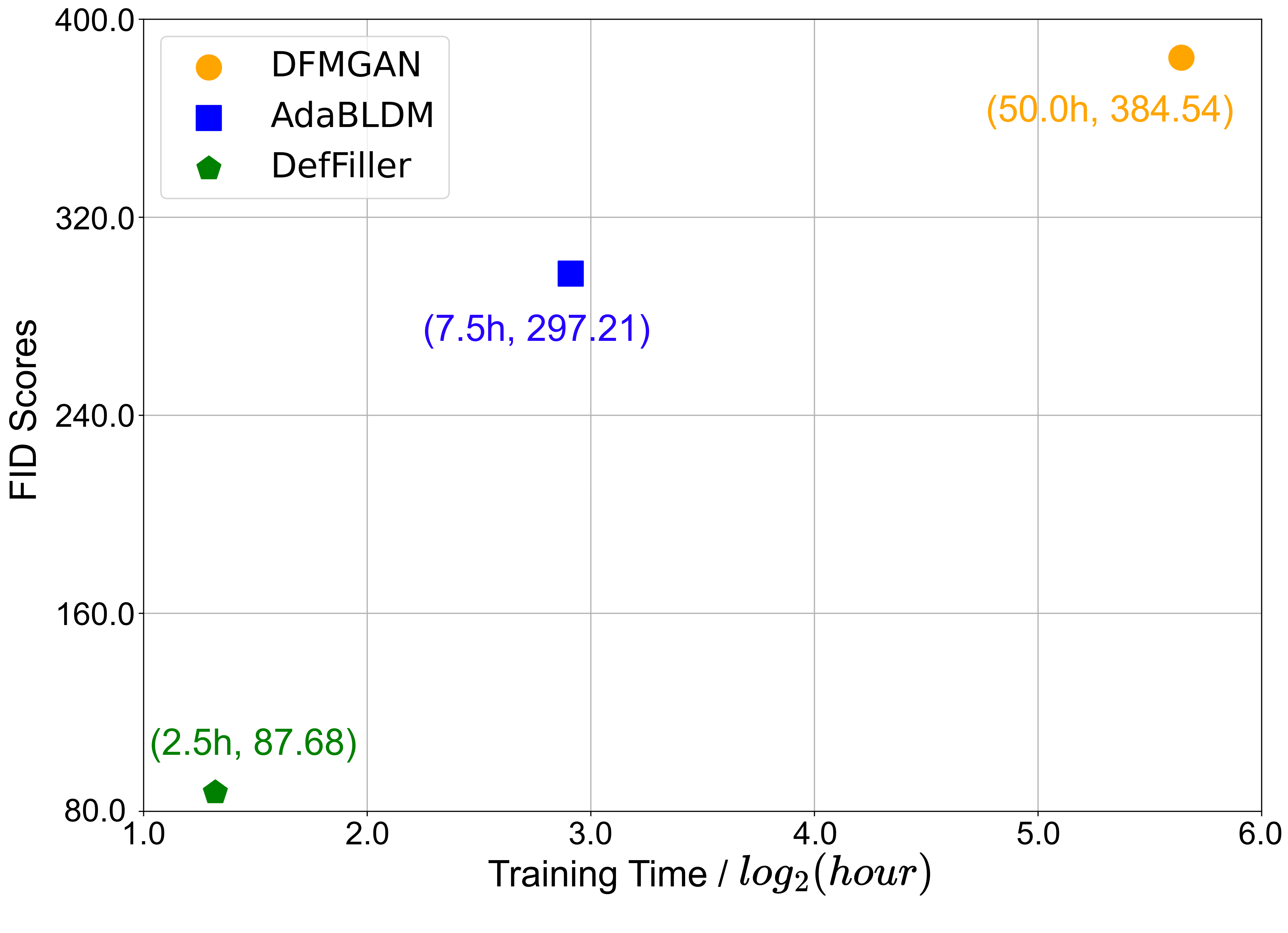}
    \caption{Comparison of generation quality and training time of DefFiller with other generative models.
    }  
    \label{speed}
\end{figure}

Additionally, we provide a thorough assessment of training efficiency and generation quality in Fig.~\ref{speed}. DFMGAN, a two-stage method, needs to first train on defect-free images and then on mask-image pairs, leading to high time costs. AdaBLDM can only generate one defect per training session, requiring multiple sessions for multi-class defect generation. In contrast, DefFiller efficiently handles multi-class defect training and produces high-quality defect samples based on text prompts and mask conditions.

\begin{table}[t]
\caption{Performance of defect detection models before and after data expansion.}
\label{comparison_expansion}
\renewcommand\arraystretch{1.2}
\begin{tabular*}{\textwidth}{@{\extracolsep\fill}lcccccc}
\toprule%
Network&Generation method&$S_\alpha \uparrow$&$\mathcal{M} \downarrow$&$E_{\xi}^{\max } \uparrow$&$F_\beta^{\max } \uparrow$\\
\midrule
    \multirow{4}*{\makecell*[c]{CSEPNet\cite{ding2022cross}}}&None&0.822&0.036&0.927&0.810\\
    &DFMGAN~\cite{Duan2023DFMGAN}&0.826&0.034&0.921&0.815\\
    &AdaBLDM~\cite{li2024novel}&0.832&0.032&0.928&0.819\\
    &DefFiller (Ours)&\textbf{0.834}&\textbf{0.030}&\textbf{0.945}&\textbf{0.830}\\
\midrule
    \multirow{4}*{\makecell*[c]{TSERNet~\cite{han2022two}}}&None&0.793&0.046&0.905&0.764\\
    &DFMGAN~\cite{Duan2023DFMGAN}&0.744&0.048&0.859&0.669\\
    &AdaBLDM~\cite{li2024novel}&0.797&0.037&0.912&0.806\\
    &DefFiller (Ours)&\textbf{0.845}&\textbf{0.031}&\textbf{0.940}&\textbf{0.842}\\
\midrule
    \multirow{4}*{\makecell*[c]{MINet~\cite{shen2024minet}}}&None&0.787&0.052&0.898&0.746\\
    &DFMGAN~\cite{Duan2023DFMGAN}&0.671&0.171&0.667&0.528\\
    &AdaBLDM~\cite{li2024novel}&0.785&0.048&0.893&0.754\\
    &DefFiller (Ours)&\textbf{0.833}&\textbf{0.033}&\textbf{0.921}&\textbf{0.816}\\
\botrule
\end{tabular*}
\end{table}

\subsubsection{Detection performance}
\label{expansion_detection}
To evaluate how the generated samples improve defect detection performance, we first split the SD-Saliency-900 dataset into a training set and a testing set with a 1:5 ratio to simulate a scarcity of defect samples. Then, we add 900 mask-image pairs created by different methods to the original training set, forming a new set. The detection models are trained on both the original and updated sets, and evaluated using the original testing set. The training parameters of detection models are the same as in Table~\ref{detection_setting}, and their performance before and after data expansion is shown in Table \ref{comparison_expansion}.
It is evident that when the number of training samples is significantly reduced, the performance of the detection model without data expansion declines noticeably. However, if the quality of the added defect samples is low, as seen with DFMGAN, they can further hinder the model's performance. In contrast, both AdaBLDM and DefFiller show consistent performance improvements when generated images are introduced, with DefFiller exhibiting a particularly strong ability to enhance steel surface defect detection. On average, the S-measure for the three detection models improves by approximately 4\%, underscoring the effectiveness and superiority of our method in expanding the dataset.

\section{Conclusion}
\label{sec5}

This paper introduces DefFiller, a mask-conditioned generation method for saliency-based defect detection. By generating high-quality defect samples with mask conditions, DefFiller eliminates the need for pixel-level annotations, allowing direct use by detection models. Using a layout-to-image diffusion model, it ensures robust synthesis even with limited data. Our thorough experiments, guided by a well-designed evaluation framework, demonstrate that DefFiller produces defect images that accurately match mask conditions and significantly enhance detection model performance. These results on the SD-Saliency-900 dataset underscore its potential for data expansion in industrial applications.

\bibliography{sn-bibliography}


\begin{thebibliography}{56}
\ifx \bisbn   \undefined \def \bisbn  #1{ISBN #1}\fi
\ifx \binits  \undefined \def \binits#1{#1}\fi
\ifx \bauthor  \undefined \def \bauthor#1{#1}\fi
\ifx \batitle  \undefined \def \batitle#1{#1}\fi
\ifx \bjtitle  \undefined \def \bjtitle#1{#1}\fi
\ifx \bvolume  \undefined \def \bvolume#1{\textbf{#1}}\fi
\ifx \byear  \undefined \def \byear#1{#1}\fi
\ifx \bissue  \undefined \def \bissue#1{#1}\fi
\ifx \bfpage  \undefined \def \bfpage#1{#1}\fi
\ifx \blpage  \undefined \def \blpage #1{#1}\fi
\ifx \burl  \undefined \def \burl#1{\textsf{#1}}\fi
\ifx \doiurl  \undefined \def \doiurl#1{\url{https://doi.org/#1}}\fi
\ifx \betal  \undefined \def \betal{\textit{et al.}}\fi
\ifx \binstitute  \undefined \def \binstitute#1{#1}\fi
\ifx \binstitutionaled  \undefined \def \binstitutionaled#1{#1}\fi
\ifx \bctitle  \undefined \def \bctitle#1{#1}\fi
\ifx \beditor  \undefined \def \beditor#1{#1}\fi
\ifx \bpublisher  \undefined \def \bpublisher#1{#1}\fi
\ifx \bbtitle  \undefined \def \bbtitle#1{#1}\fi
\ifx \bedition  \undefined \def \bedition#1{#1}\fi
\ifx \bseriesno  \undefined \def \bseriesno#1{#1}\fi
\ifx \blocation  \undefined \def \blocation#1{#1}\fi
\ifx \bsertitle  \undefined \def \bsertitle#1{#1}\fi
\ifx \bsnm \undefined \def \bsnm#1{#1}\fi
\ifx \bsuffix \undefined \def \bsuffix#1{#1}\fi
\ifx \bparticle \undefined \def \bparticle#1{#1}\fi
\ifx \barticle \undefined \def \barticle#1{#1}\fi
\bibcommenthead
\ifx \bconfdate \undefined \def \bconfdate #1{#1}\fi
\ifx \botherref \undefined \def \botherref #1{#1}\fi
\ifx \url \undefined \def \url#1{\textsf{#1}}\fi
\ifx \bchapter \undefined \def \bchapter#1{#1}\fi
\ifx \bbook \undefined \def \bbook#1{#1}\fi
\ifx \bcomment \undefined \def \bcomment#1{#1}\fi
\ifx \oauthor \undefined \def \oauthor#1{#1}\fi
\ifx \citeauthoryear \undefined \def \citeauthoryear#1{#1}\fi
\ifx \endbibitem  \undefined \def \endbibitem {}\fi
\ifx \bconflocation  \undefined \def \bconflocation#1{#1}\fi
\ifx \arxivurl  \undefined \def \arxivurl#1{\textsf{#1}}\fi
\csname PreBibitemsHook\endcsname

\bibitem[\protect\citeauthoryear{Wu and He}{2024}]{wu2024efficient}
\begin{botherref}
\oauthor{\bsnm{Wu}, \binits{C.}},
\oauthor{\bsnm{He}, \binits{T.}}:
Efficient minor defects detection on steel surface via res-attention and position encoding.
The Visual Computer,
1--15
(2024)
\end{botherref}
\endbibitem

\bibitem[\protect\citeauthoryear{Zhang and Tian}{2023}]{zhang2023transformer}
\begin{barticle}
\bauthor{\bsnm{Zhang}, \binits{M.}},
\bauthor{\bsnm{Tian}, \binits{X.}}:
\batitle{Transformer architecture based on mutual attention for image-anomaly detection}.
\bjtitle{Virtual Reality \& Intelligent Hardware}
\bvolume{5}(\bissue{1}),
\bfpage{57}--\blpage{67}
(\byear{2023})
\end{barticle}
\endbibitem

\bibitem[\protect\citeauthoryear{Tian et~al.}{2023}]{tian2023ilidviz}
\begin{barticle}
\bauthor{\bsnm{Tian}, \binits{X.}},
\bauthor{\bsnm{Wu}, \binits{Z.}},
\bauthor{\bsnm{Cao}, \binits{J.}},
\bauthor{\bsnm{Chen}, \binits{S.}},
\bauthor{\bsnm{Dong}, \binits{X.}}:
\batitle{Ilidviz: An incremental learning-based visual analysis system for network anomaly detection}.
\bjtitle{Virtual Reality \& Intelligent Hardware}
\bvolume{5}(\bissue{6}),
\bfpage{471}--\blpage{489}
(\byear{2023})
\end{barticle}
\endbibitem

\bibitem[\protect\citeauthoryear{Sun et~al.}{2024}]{sun2024adversarial}
\begin{botherref}
\oauthor{\bsnm{Sun}, \binits{W.}},
\oauthor{\bsnm{Zhang}, \binits{J.}},
\oauthor{\bsnm{Liu}, \binits{Y.}}:
Adversarial-based refinement dual-branch network for semi-supervised salient object detection of strip steel surface defects.
The Visual Computer,
1--15
(2024)
\end{botherref}
\endbibitem

\bibitem[\protect\citeauthoryear{Ding et~al.}{2022}]{ding2022cross}
\begin{barticle}
\bauthor{\bsnm{Ding}, \binits{T.}},
\bauthor{\bsnm{Li}, \binits{G.}},
\bauthor{\bsnm{Liu}, \binits{Z.}},
\bauthor{\bsnm{Wang}, \binits{Y.}}:
\batitle{Cross-scale edge purification network for salient object detection of steel defect images}.
\bjtitle{Measurement}
\bvolume{199},
\bfpage{111429}
(\byear{2022})
\end{barticle}
\endbibitem

\bibitem[\protect\citeauthoryear{Han et~al.}{2022}]{han2022two}
\begin{barticle}
\bauthor{\bsnm{Han}, \binits{C.}},
\bauthor{\bsnm{Li}, \binits{G.}},
\bauthor{\bsnm{Liu}, \binits{Z.}}:
\batitle{Two-stage edge reuse network for salient object detection of strip steel surface defects}.
\bjtitle{IEEE Transactions on Instrumentation and Measurement}
\bvolume{71},
\bfpage{1}--\blpage{12}
(\byear{2022})
\end{barticle}
\endbibitem

\bibitem[\protect\citeauthoryear{Shen et~al.}{2024}]{shen2024minet}
\begin{botherref}
\oauthor{\bsnm{Shen}, \binits{K.}},
\oauthor{\bsnm{Zhou}, \binits{X.}},
\oauthor{\bsnm{Liu}, \binits{Z.}}:
Minet: Multiscale interactive network for real-time salient object detection of strip steel surface defects.
IEEE Transactions on Industrial Informatics
(2024)
\end{botherref}
\endbibitem

\bibitem[\protect\citeauthoryear{Wan et~al.}{2023}]{wan2023lfrnet}
\begin{barticle}
\bauthor{\bsnm{Wan}, \binits{B.}},
\bauthor{\bsnm{Zhou}, \binits{X.}},
\bauthor{\bsnm{Zheng}, \binits{B.}},
\bauthor{\bsnm{Yin}, \binits{H.}},
\bauthor{\bsnm{Zhu}, \binits{Z.}},
\bauthor{\bsnm{Wang}, \binits{H.}},
\bauthor{\bsnm{Sun}, \binits{Y.}},
\bauthor{\bsnm{Zhang}, \binits{J.}},
\bauthor{\bsnm{Yan}, \binits{C.}}:
\batitle{Lfrnet: Localizing, focus, and refinement network for salient object detection of surface defects}.
\bjtitle{IEEE Transactions on Instrumentation and Measurement}
\bvolume{72},
\bfpage{1}--\blpage{12}
(\byear{2023})
\end{barticle}
\endbibitem

\bibitem[\protect\citeauthoryear{Wei et~al.}{2023}]{wei2023diversified}
\begin{bchapter}
\bauthor{\bsnm{Wei}, \binits{J.}},
\bauthor{\bsnm{Shen}, \binits{F.}},
\bauthor{\bsnm{Lv}, \binits{C.}},
\bauthor{\bsnm{Zhang}, \binits{Z.}},
\bauthor{\bsnm{Zhang}, \binits{F.}},
\bauthor{\bsnm{Yang}, \binits{H.}}:
\bctitle{Diversified and multi-class controllable industrial defect synthesis for data augmentation and transfer}.
In: \bbtitle{Proceedings of the IEEE/CVF Conference on Computer Vision and Pattern Recognition},
pp. \bfpage{4444}--\blpage{4452}
(\byear{2023})
\end{bchapter}
\endbibitem

\bibitem[\protect\citeauthoryear{Duan et~al.}{2023}]{Duan2023DFMGAN}
\begin{bchapter}
\bauthor{\bsnm{Duan}, \binits{Y.}},
\bauthor{\bsnm{Hong}, \binits{Y.}},
\bauthor{\bsnm{Niu}, \binits{L.}},
\bauthor{\bsnm{Zhang}, \binits{L.}}:
\bctitle{Few-shot defect image generation via defect-aware feature manipulation}.
In: \bbtitle{AAAI}
(\byear{2023})
\end{bchapter}
\endbibitem

\bibitem[\protect\citeauthoryear{Li et~al.}{2024}]{li2024novel}
\begin{botherref}
\oauthor{\bsnm{Li}, \binits{H.}},
\oauthor{\bsnm{Zhang}, \binits{Z.}},
\oauthor{\bsnm{Chen}, \binits{H.}},
\oauthor{\bsnm{Wu}, \binits{L.}},
\oauthor{\bsnm{Li}, \binits{B.}},
\oauthor{\bsnm{Liu}, \binits{D.}},
\oauthor{\bsnm{Wang}, \binits{M.}}:
A novel approach to industrial defect generation through blended latent diffusion model with online adaptation.
arXiv preprint arXiv:2402.19330
(2024)
\end{botherref}
\endbibitem

\bibitem[\protect\citeauthoryear{Ho et~al.}{2020}]{ho2020denoising}
\begin{barticle}
\bauthor{\bsnm{Ho}, \binits{J.}},
\bauthor{\bsnm{Jain}, \binits{A.}},
\bauthor{\bsnm{Abbeel}, \binits{P.}}:
\batitle{Denoising diffusion probabilistic models}.
\bjtitle{Advances in neural information processing systems}
\bvolume{33},
\bfpage{6840}--\blpage{6851}
(\byear{2020})
\end{barticle}
\endbibitem

\bibitem[\protect\citeauthoryear{Rombach et~al.}{2022}]{rombach2022high}
\begin{bchapter}
\bauthor{\bsnm{Rombach}, \binits{R.}},
\bauthor{\bsnm{Blattmann}, \binits{A.}},
\bauthor{\bsnm{Lorenz}, \binits{D.}},
\bauthor{\bsnm{Esser}, \binits{P.}},
\bauthor{\bsnm{Ommer}, \binits{B.}}:
\bctitle{High-resolution image synthesis with latent diffusion models}.
In: \bbtitle{Proceedings of the IEEE/CVF Conference on Computer Vision and Pattern Recognition},
pp. \bfpage{10684}--\blpage{10695}
(\byear{2022})
\end{bchapter}
\endbibitem

\bibitem[\protect\citeauthoryear{Balaji et~al.}{2022}]{balaji2022ediffi}
\begin{botherref}
\oauthor{\bsnm{Balaji}, \binits{Y.}},
\oauthor{\bsnm{Nah}, \binits{S.}},
\oauthor{\bsnm{Huang}, \binits{X.}},
\oauthor{\bsnm{Vahdat}, \binits{A.}},
\oauthor{\bsnm{Song}, \binits{J.}},
\oauthor{\bsnm{Kreis}, \binits{K.}},
\oauthor{\bsnm{Aittala}, \binits{M.}},
\oauthor{\bsnm{Aila}, \binits{T.}},
\oauthor{\bsnm{Laine}, \binits{S.}},
\oauthor{\bsnm{Catanzaro}, \binits{B.}}, et al.:
ediffi: Text-to-image diffusion models with an ensemble of expert denoisers.
arXiv preprint arXiv:2211.01324
(2022)
\end{botherref}
\endbibitem

\bibitem[\protect\citeauthoryear{Wu et~al.}{2024}]{wu2024ddfa}
\begin{botherref}
\oauthor{\bsnm{Wu}, \binits{H.}},
\oauthor{\bsnm{Li}, \binits{B.}},
\oauthor{\bsnm{Tian}, \binits{L.}},
\oauthor{\bsnm{Dong}, \binits{C.}}:
Ddfa: a displacement and diffusion-based feature augmentation method for imbalanced image recognition.
The Visual Computer,
1--15
(2024)
\end{botherref}
\endbibitem

\bibitem[\protect\citeauthoryear{Yang et~al.}{2024}]{yang2024novel}
\begin{botherref}
\oauthor{\bsnm{Yang}, \binits{X.}},
\oauthor{\bsnm{Ye}, \binits{T.}},
\oauthor{\bsnm{Yuan}, \binits{X.}},
\oauthor{\bsnm{Zhu}, \binits{W.}},
\oauthor{\bsnm{Mei}, \binits{X.}},
\oauthor{\bsnm{Zhou}, \binits{F.}}:
A novel data augmentation method based on denoising diffusion probabilistic model for fault diagnosis under imbalanced data.
IEEE Transactions on Industrial Informatics
(2024)
\end{botherref}
\endbibitem

\bibitem[\protect\citeauthoryear{Tai et~al.}{2024}]{tai2024defect}
\begin{botherref}
\oauthor{\bsnm{Tai}, \binits{Y.}},
\oauthor{\bsnm{Yang}, \binits{K.}},
\oauthor{\bsnm{Peng}, \binits{T.}},
\oauthor{\bsnm{Huang}, \binits{Z.}},
\oauthor{\bsnm{Zhang}, \binits{Z.}}:
Defect image sample generation with diffusion prior for steel surface defect recognition.
IEEE Transactions on Automation Science and Engineering,
1--13
(2024)
\end{botherref}
\endbibitem

\bibitem[\protect\citeauthoryear{Chen et~al.}{2024}]{chen2024training}
\begin{bchapter}
\bauthor{\bsnm{Chen}, \binits{M.}},
\bauthor{\bsnm{Laina}, \binits{I.}},
\bauthor{\bsnm{Vedaldi}, \binits{A.}}:
\bctitle{Training-free layout control with cross-attention guidance}.
In: \bbtitle{Proceedings of the IEEE/CVF Winter Conference on Applications of Computer Vision},
pp. \bfpage{5343}--\blpage{5353}
(\byear{2024})
\end{bchapter}
\endbibitem

\bibitem[\protect\citeauthoryear{Xie et~al.}{2023}]{xie2023boxdiff}
\begin{bchapter}
\bauthor{\bsnm{Xie}, \binits{J.}},
\bauthor{\bsnm{Li}, \binits{Y.}},
\bauthor{\bsnm{Huang}, \binits{Y.}},
\bauthor{\bsnm{Liu}, \binits{H.}},
\bauthor{\bsnm{Zhang}, \binits{W.}},
\bauthor{\bsnm{Zheng}, \binits{Y.}},
\bauthor{\bsnm{Shou}, \binits{M.Z.}}:
\bctitle{Boxdiff: Text-to-image synthesis with training-free box-constrained diffusion}.
In: \bbtitle{Proceedings of the IEEE/CVF International Conference on Computer Vision},
pp. \bfpage{7452}--\blpage{7461}
(\byear{2023})
\end{bchapter}
\endbibitem

\bibitem[\protect\citeauthoryear{Li et~al.}{2023}]{li2023gligen}
\begin{botherref}
\oauthor{\bsnm{Li}, \binits{Y.}},
\oauthor{\bsnm{Liu}, \binits{H.}},
\oauthor{\bsnm{Wu}, \binits{Q.}},
\oauthor{\bsnm{Mu}, \binits{F.}},
\oauthor{\bsnm{Yang}, \binits{J.}},
\oauthor{\bsnm{Gao}, \binits{J.}},
\oauthor{\bsnm{Li}, \binits{C.}},
\oauthor{\bsnm{Lee}, \binits{Y.J.}}:
Gligen: Open-set grounded text-to-image generation.
CVPR
(2023)
\end{botherref}
\endbibitem

\bibitem[\protect\citeauthoryear{Wang et~al.}{2024}]{Wang_2024_CVPR}
\begin{bchapter}
\bauthor{\bsnm{Wang}, \binits{X.}},
\bauthor{\bsnm{Darrell}, \binits{T.}},
\bauthor{\bsnm{Rambhatla}, \binits{S.S.}},
\bauthor{\bsnm{Girdhar}, \binits{R.}},
\bauthor{\bsnm{Misra}, \binits{I.}}:
\bctitle{Instancediffusion: Instance-level control for image generation}.
In: \bbtitle{Proceedings of the IEEE/CVF Conference on Computer Vision and Pattern Recognition (CVPR)},
pp. \bfpage{6232}--\blpage{6242}
(\byear{2024})
\end{bchapter}
\endbibitem

\bibitem[\protect\citeauthoryear{Endo}{2024}]{endo2024masked}
\begin{barticle}
\bauthor{\bsnm{Endo}, \binits{Y.}}:
\batitle{Masked-attention diffusion guidance for spatially controlling text-to-image generation}.
\bjtitle{The Visual Computer}
\bvolume{40}(\bissue{9}),
\bfpage{6033}--\blpage{6045}
(\byear{2024})
\end{barticle}
\endbibitem

\bibitem[\protect\citeauthoryear{Goodfellow et~al.}{2014}]{goodfellow2014generative}
\begin{botherref}
\oauthor{\bsnm{Goodfellow}, \binits{I.}},
\oauthor{\bsnm{Pouget-Abadie}, \binits{J.}},
\oauthor{\bsnm{Mirza}, \binits{M.}},
\oauthor{\bsnm{Xu}, \binits{B.}},
\oauthor{\bsnm{Warde-Farley}, \binits{D.}},
\oauthor{\bsnm{Ozair}, \binits{S.}},
\oauthor{\bsnm{Courville}, \binits{A.}},
\oauthor{\bsnm{Bengio}, \binits{Y.}}:
Generative adversarial nets.
Advances in neural information processing systems
\textbf{27}
(2014)
\end{botherref}
\endbibitem

\bibitem[\protect\citeauthoryear{Zhu et~al.}{2017}]{zhu2017unpaired}
\begin{bchapter}
\bauthor{\bsnm{Zhu}, \binits{J.-Y.}},
\bauthor{\bsnm{Park}, \binits{T.}},
\bauthor{\bsnm{Isola}, \binits{P.}},
\bauthor{\bsnm{Efros}, \binits{A.A.}}:
\bctitle{Unpaired image-to-image translation using cycle-consistent adversarial networks}.
In: \bbtitle{Proceedings of the IEEE International Conference on Computer Vision},
pp. \bfpage{2223}--\blpage{2232}
(\byear{2017})
\end{bchapter}
\endbibitem

\bibitem[\protect\citeauthoryear{Sauer et~al.}{2022}]{sauer2022stylegan}
\begin{bchapter}
\bauthor{\bsnm{Sauer}, \binits{A.}},
\bauthor{\bsnm{Schwarz}, \binits{K.}},
\bauthor{\bsnm{Geiger}, \binits{A.}}:
\bctitle{Stylegan-xl: Scaling stylegan to large diverse datasets}.
In: \bbtitle{ACM SIGGRAPH 2022 Conference Proceedings},
pp. \bfpage{1}--\blpage{10}
(\byear{2022})
\end{bchapter}
\endbibitem

\bibitem[\protect\citeauthoryear{Zhao et~al.}{2024}]{zhao2024gan}
\begin{barticle}
\bauthor{\bsnm{Zhao}, \binits{W.}},
\bauthor{\bsnm{Zhu}, \binits{J.}},
\bauthor{\bsnm{Huang}, \binits{J.}},
\bauthor{\bsnm{Li}, \binits{P.}},
\bauthor{\bsnm{Sheng}, \binits{B.}}:
\batitle{Gan-based multi-decomposition photo cartoonization}.
\bjtitle{Computer Animation and Virtual Worlds}
\bvolume{35}(\bissue{3}),
\bfpage{2248}
(\byear{2024})
\end{barticle}
\endbibitem

\bibitem[\protect\citeauthoryear{Hu et~al.}{2024}]{hu2024msembgan}
\begin{botherref}
\oauthor{\bsnm{Hu}, \binits{X.}},
\oauthor{\bsnm{Yang}, \binits{C.}},
\oauthor{\bsnm{Fang}, \binits{F.}},
\oauthor{\bsnm{Huang}, \binits{J.}},
\oauthor{\bsnm{Li}, \binits{P.}},
\oauthor{\bsnm{ShengB}, \binits{B.}},
\oauthor{\bsnm{Lee}, \binits{T.-Y.}}:
Msembgan: Multi-stitch embroidery synthesis via region-aware texture generation.
IEEE Transactions on Visualization and Computer Graphics
(2024)
\end{botherref}
\endbibitem

\bibitem[\protect\citeauthoryear{Zhang et~al.}{2021}]{zhang2021defect}
\begin{bchapter}
\bauthor{\bsnm{Zhang}, \binits{G.}},
\bauthor{\bsnm{Cui}, \binits{K.}},
\bauthor{\bsnm{Hung}, \binits{T.-Y.}},
\bauthor{\bsnm{Lu}, \binits{S.}}:
\bctitle{Defect-gan: High-fidelity defect synthesis for automated defect inspection}.
In: \bbtitle{Proceedings of the IEEE/CVF Winter Conference on Applications of Computer Vision},
pp. \bfpage{2524}--\blpage{2534}
(\byear{2021})
\end{bchapter}
\endbibitem

\bibitem[\protect\citeauthoryear{Zhao et~al.}{2023}]{zhao2023defect}
\begin{botherref}
\oauthor{\bsnm{Zhao}, \binits{C.}},
\oauthor{\bsnm{Xue}, \binits{W.}},
\oauthor{\bsnm{Fu}, \binits{W.}},
\oauthor{\bsnm{Li}, \binits{Z.}},
\oauthor{\bsnm{Fang}, \binits{X.}}:
Defect sample image generation method based on gans in diamond tool defect detection.
IEEE Transactions on Instrumentation and Measurement
(2023)
\end{botherref}
\endbibitem

\bibitem[\protect\citeauthoryear{Li et~al.}{2023}]{li2023dls}
\begin{botherref}
\oauthor{\bsnm{Li}, \binits{W.}},
\oauthor{\bsnm{Gu}, \binits{C.}},
\oauthor{\bsnm{Chen}, \binits{J.}},
\oauthor{\bsnm{Ma}, \binits{C.}},
\oauthor{\bsnm{Zhang}, \binits{X.}},
\oauthor{\bsnm{Chen}, \binits{B.}},
\oauthor{\bsnm{Wan}, \binits{S.}}:
Dls-gan: generative adversarial nets for defect location sensitive data augmentation.
IEEE Transactions on Automation Science and Engineering
(2023)
\end{botherref}
\endbibitem

\bibitem[\protect\citeauthoryear{Ran et~al.}{2024}]{ran2024sketch}
\begin{barticle}
\bauthor{\bsnm{Ran}, \binits{G.}},
\bauthor{\bsnm{Yao}, \binits{X.}},
\bauthor{\bsnm{Wang}, \binits{K.}},
\bauthor{\bsnm{Ye}, \binits{J.}},
\bauthor{\bsnm{Ou}, \binits{S.}}:
\batitle{Sketch-guided spatial adaptive normalization and high-level feature constraints based gan image synthesis for steel strip defect detection data augmentation}.
\bjtitle{Measurement Science and Technology}
\bvolume{35}(\bissue{4}),
\bfpage{045408}
(\byear{2024})
\end{barticle}
\endbibitem

\bibitem[\protect\citeauthoryear{Ramesh et~al.}{2022}]{ramesh2022hierarchical}
\begin{barticle}
\bauthor{\bsnm{Ramesh}, \binits{A.}},
\bauthor{\bsnm{Dhariwal}, \binits{P.}},
\bauthor{\bsnm{Nichol}, \binits{A.}},
\bauthor{\bsnm{Chu}, \binits{C.}},
\bauthor{\bsnm{Chen}, \binits{M.}}:
\batitle{Hierarchical text-conditional image generation with clip latents}.
\bjtitle{arXiv preprint arXiv:2204.06125}
\bvolume{1}(\bissue{2}),
\bfpage{3}
(\byear{2022})
\end{barticle}
\endbibitem

\bibitem[\protect\citeauthoryear{Ma et~al.}{2024}]{ma2024diffspeaker}
\begin{botherref}
\oauthor{\bsnm{Ma}, \binits{Z.}},
\oauthor{\bsnm{Zhu}, \binits{X.}},
\oauthor{\bsnm{Qi}, \binits{G.}},
\oauthor{\bsnm{Qian}, \binits{C.}},
\oauthor{\bsnm{Zhang}, \binits{Z.}},
\oauthor{\bsnm{Lei}, \binits{Z.}}:
Diffspeaker: Speech-driven 3d facial animation with diffusion transformer.
arXiv preprint arXiv:2402.05712
(2024)
\end{botherref}
\endbibitem

\bibitem[\protect\citeauthoryear{Ma et~al.}{2025}]{ma2025scaledreamer}
\begin{bchapter}
\bauthor{\bsnm{Ma}, \binits{Z.}},
\bauthor{\bsnm{Wei}, \binits{Y.}},
\bauthor{\bsnm{Zhang}, \binits{Y.}},
\bauthor{\bsnm{Zhu}, \binits{X.}},
\bauthor{\bsnm{Lei}, \binits{Z.}},
\bauthor{\bsnm{Zhang}, \binits{L.}}:
\bctitle{Scaledreamer: Scalable text-to-3d synthesis with asynchronous score distillation}.
In: \bbtitle{European Conference on Computer Vision},
pp. \bfpage{1}--\blpage{19}
(\byear{2025}).
\bcomment{Springer}
\end{bchapter}
\endbibitem

\bibitem[\protect\citeauthoryear{Wu et~al.}{2024}]{wu2024one}
\begin{botherref}
\oauthor{\bsnm{Wu}, \binits{R.}},
\oauthor{\bsnm{Sun}, \binits{L.}},
\oauthor{\bsnm{Ma}, \binits{Z.}},
\oauthor{\bsnm{Zhang}, \binits{L.}}:
One-step effective diffusion network for real-world image super-resolution.
arXiv preprint arXiv:2406.08177
(2024)
\end{botherref}
\endbibitem

\bibitem[\protect\citeauthoryear{Xiao et~al.}{2024}]{xiao2024parameter}
\begin{barticle}
\bauthor{\bsnm{Xiao}, \binits{Z.}},
\bauthor{\bsnm{Li}, \binits{C.}},
\bauthor{\bsnm{Liu}, \binits{T.}},
\bauthor{\bsnm{Liu}, \binits{W.}},
\bauthor{\bsnm{Mo}, \binits{S.}},
\bauthor{\bsnm{Houjoh}, \binits{H.}}:
\batitle{Parameter sharing fault data generation method based on diffusion model under imbalance data}.
\bjtitle{Measurement Science and Technology}
\bvolume{35}(\bissue{10}),
\bfpage{106111}
(\byear{2024})
\end{barticle}
\endbibitem

\bibitem[\protect\citeauthoryear{Zhang et~al.}{2024}]{zhang2024ag}
\begin{barticle}
\bauthor{\bsnm{Zhang}, \binits{M.}},
\bauthor{\bsnm{Yang}, \binits{J.}},
\bauthor{\bsnm{Xian}, \binits{Y.}},
\bauthor{\bsnm{Li}, \binits{W.}},
\bauthor{\bsnm{Gu}, \binits{J.}},
\bauthor{\bsnm{Meng}, \binits{W.}},
\bauthor{\bsnm{Zhang}, \binits{J.}},
\bauthor{\bsnm{Zhang}, \binits{X.}}:
\batitle{Ag-sdm: Aquascape generation based on stable diffusion model with low-rank adaptation}.
\bjtitle{Computer Animation and Virtual Worlds}
\bvolume{35}(\bissue{3}),
\bfpage{2252}
(\byear{2024})
\end{barticle}
\endbibitem

\bibitem[\protect\citeauthoryear{Zhang et~al.}{2023}]{zhang2023adding}
\begin{bchapter}
\bauthor{\bsnm{Zhang}, \binits{L.}},
\bauthor{\bsnm{Rao}, \binits{A.}},
\bauthor{\bsnm{Agrawala}, \binits{M.}}:
\bctitle{Adding conditional control to text-to-image diffusion models}.
In: \bbtitle{Proceedings of the IEEE/CVF International Conference on Computer Vision},
pp. \bfpage{3836}--\blpage{3847}
(\byear{2023})
\end{bchapter}
\endbibitem

\bibitem[\protect\citeauthoryear{Avrahami et~al.}{2023}]{avrahami2023blended}
\begin{barticle}
\bauthor{\bsnm{Avrahami}, \binits{O.}},
\bauthor{\bsnm{Fried}, \binits{O.}},
\bauthor{\bsnm{Lischinski}, \binits{D.}}:
\batitle{Blended latent diffusion}.
\bjtitle{ACM transactions on graphics (TOG)}
\bvolume{42}(\bissue{4}),
\bfpage{1}--\blpage{11}
(\byear{2023})
\end{barticle}
\endbibitem

\bibitem[\protect\citeauthoryear{Achanta et~al.}{2009}]{achanta2009frequency}
\begin{bchapter}
\bauthor{\bsnm{Achanta}, \binits{R.}},
\bauthor{\bsnm{Hemami}, \binits{S.}},
\bauthor{\bsnm{Estrada}, \binits{F.}},
\bauthor{\bsnm{Susstrunk}, \binits{S.}}:
\bctitle{Frequency-tuned salient region detection}.
In: \bbtitle{2009 IEEE Conference on Computer Vision and Pattern Recognition},
pp. \bfpage{1597}--\blpage{1604}
(\byear{2009}).
\bcomment{IEEE}
\end{bchapter}
\endbibitem

\bibitem[\protect\citeauthoryear{Ma et~al.}{2023}]{ma2023skin}
\begin{barticle}
\bauthor{\bsnm{Ma}, \binits{C.}},
\bauthor{\bsnm{He}, \binits{T.}},
\bauthor{\bsnm{Gao}, \binits{J.}}:
\batitle{Skin scar segmentation based on saliency detection}.
\bjtitle{The Visual Computer}
\bvolume{39}(\bissue{10}),
\bfpage{4887}--\blpage{4899}
(\byear{2023})
\end{barticle}
\endbibitem

\bibitem[\protect\citeauthoryear{Zhou et~al.}{2021a}]{zhou2021dense}
\begin{barticle}
\bauthor{\bsnm{Zhou}, \binits{X.}},
\bauthor{\bsnm{Fang}, \binits{H.}},
\bauthor{\bsnm{Liu}, \binits{Z.}},
\bauthor{\bsnm{Zheng}, \binits{B.}},
\bauthor{\bsnm{Sun}, \binits{Y.}},
\bauthor{\bsnm{Zhang}, \binits{J.}},
\bauthor{\bsnm{Yan}, \binits{C.}}:
\batitle{Dense attention-guided cascaded network for salient object detection of strip steel surface defects}.
\bjtitle{IEEE Transactions on Instrumentation and Measurement}
\bvolume{71},
\bfpage{1}--\blpage{14}
(\byear{2021})
\end{barticle}
\endbibitem

\bibitem[\protect\citeauthoryear{Zhou et~al.}{2021b}]{zhou2021edge}
\begin{barticle}
\bauthor{\bsnm{Zhou}, \binits{X.}},
\bauthor{\bsnm{Fang}, \binits{H.}},
\bauthor{\bsnm{Fei}, \binits{X.}},
\bauthor{\bsnm{Shi}, \binits{R.}},
\bauthor{\bsnm{Zhang}, \binits{J.}}:
\batitle{Edge-aware multi-level interactive network for salient object detection of strip steel surface defects}.
\bjtitle{IEEE Access}
\bvolume{9},
\bfpage{149465}--\blpage{149476}
(\byear{2021})
\end{barticle}
\endbibitem

\bibitem[\protect\citeauthoryear{Dong et~al.}{2019}]{dong2019pga}
\begin{barticle}
\bauthor{\bsnm{Dong}, \binits{H.}},
\bauthor{\bsnm{Song}, \binits{K.}},
\bauthor{\bsnm{He}, \binits{Y.}},
\bauthor{\bsnm{Xu}, \binits{J.}},
\bauthor{\bsnm{Yan}, \binits{Y.}},
\bauthor{\bsnm{Meng}, \binits{Q.}}:
\batitle{Pga-net: Pyramid feature fusion and global context attention network for automated surface defect detection}.
\bjtitle{IEEE Transactions on Industrial Informatics}
\bvolume{16}(\bissue{12}),
\bfpage{7448}--\blpage{7458}
(\byear{2019})
\end{barticle}
\endbibitem

\bibitem[\protect\citeauthoryear{Wan et~al.}{2023}]{wan2023sminet}
\begin{barticle}
\bauthor{\bsnm{Wan}, \binits{B.}},
\bauthor{\bsnm{Zhou}, \binits{X.}},
\bauthor{\bsnm{Sun}, \binits{Y.}},
\bauthor{\bsnm{Zhu}, \binits{Z.}},
\bauthor{\bsnm{Yin}, \binits{H.}},
\bauthor{\bsnm{Hu}, \binits{J.}},
\bauthor{\bsnm{Zhang}, \binits{J.}},
\bauthor{\bsnm{Yan}, \binits{C.}}:
\batitle{Sminet: Semantics-aware multi-level feature interaction network for surface defect detection}.
\bjtitle{Engineering Applications of Artificial Intelligence}
\bvolume{123},
\bfpage{106474}
(\byear{2023})
\end{barticle}
\endbibitem

\bibitem[\protect\citeauthoryear{Liu et~al.}{2022}]{liu2022convnet}
\begin{botherref}
\oauthor{\bsnm{Liu}, \binits{Z.}},
\oauthor{\bsnm{Mao}, \binits{H.}},
\oauthor{\bsnm{Wu}, \binits{C.-Y.}},
\oauthor{\bsnm{Feichtenhofer}, \binits{C.}},
\oauthor{\bsnm{Darrell}, \binits{T.}},
\oauthor{\bsnm{Xie}, \binits{S.}}:
A convnet for the 2020s.
Proceedings of the IEEE/CVF Conference on Computer Vision and Pattern Recognition (CVPR)
(2022)
\end{botherref}
\endbibitem

\bibitem[\protect\citeauthoryear{Radford et~al.}{2021}]{radford2021learning}
\begin{bchapter}
\bauthor{\bsnm{Radford}, \binits{A.}},
\bauthor{\bsnm{Kim}, \binits{J.W.}},
\bauthor{\bsnm{Hallacy}, \binits{C.}},
\bauthor{\bsnm{Ramesh}, \binits{A.}},
\bauthor{\bsnm{Goh}, \binits{G.}},
\bauthor{\bsnm{Agarwal}, \binits{S.}},
\bauthor{\bsnm{Sastry}, \binits{G.}},
\bauthor{\bsnm{Askell}, \binits{A.}},
\bauthor{\bsnm{Mishkin}, \binits{P.}},
\bauthor{\bsnm{Clark}, \binits{J.}}, \betal:
\bctitle{Learning transferable visual models from natural language supervision}.
In: \bbtitle{International Conference on Machine Learning},
pp. \bfpage{8748}--\blpage{8763}
(\byear{2021}).
\bcomment{PMLR}
\end{bchapter}
\endbibitem

\bibitem[\protect\citeauthoryear{Ho and Salimans}{2022}]{ho2022classifier}
\begin{botherref}
\oauthor{\bsnm{Ho}, \binits{J.}},
\oauthor{\bsnm{Salimans}, \binits{T.}}:
Classifier-free diffusion guidance.
arXiv preprint arXiv:2207.12598
(2022)
\end{botherref}
\endbibitem

\bibitem[\protect\citeauthoryear{Heusel et~al.}{2017}]{heusel2017gans}
\begin{botherref}
\oauthor{\bsnm{Heusel}, \binits{M.}},
\oauthor{\bsnm{Ramsauer}, \binits{H.}},
\oauthor{\bsnm{Unterthiner}, \binits{T.}},
\oauthor{\bsnm{Nessler}, \binits{B.}},
\oauthor{\bsnm{Hochreiter}, \binits{S.}}:
Gans trained by a two time-scale update rule converge to a local nash equilibrium.
Advances in neural information processing systems
\textbf{30}
(2017)
\end{botherref}
\endbibitem

\bibitem[\protect\citeauthoryear{Fan et~al.}{2017}]{fan2017structure}
\begin{bchapter}
\bauthor{\bsnm{Fan}, \binits{D.-P.}},
\bauthor{\bsnm{Cheng}, \binits{M.-M.}},
\bauthor{\bsnm{Liu}, \binits{Y.}},
\bauthor{\bsnm{Li}, \binits{T.}},
\bauthor{\bsnm{Borji}, \binits{A.}}:
\bctitle{Structure-measure: A new way to evaluate foreground maps}.
In: \bbtitle{Proceedings of the IEEE International Conference on Computer Vision},
pp. \bfpage{4548}--\blpage{4557}
(\byear{2017})
\end{bchapter}
\endbibitem

\bibitem[\protect\citeauthoryear{Fan et~al.}{2018}]{fan2018enhanced}
\begin{botherref}
\oauthor{\bsnm{Fan}, \binits{D.-P.}},
\oauthor{\bsnm{Gong}, \binits{C.}},
\oauthor{\bsnm{Cao}, \binits{Y.}},
\oauthor{\bsnm{Ren}, \binits{B.}},
\oauthor{\bsnm{Cheng}, \binits{M.-M.}},
\oauthor{\bsnm{Borji}, \binits{A.}}:
Enhanced-alignment measure for binary foreground map evaluation.
arXiv preprint arXiv:1805.10421
(2018)
\end{botherref}
\endbibitem

\bibitem[\protect\citeauthoryear{Song et~al.}{2020}]{song2020edrnet}
\begin{barticle}
\bauthor{\bsnm{Song}, \binits{G.}},
\bauthor{\bsnm{Song}, \binits{K.}},
\bauthor{\bsnm{Yan}, \binits{Y.}}:
\batitle{Edrnet: Encoder--decoder residual network for salient object detection of strip steel surface defects}.
\bjtitle{IEEE Transactions on Instrumentation and Measurement}
\bvolume{69}(\bissue{12}),
\bfpage{9709}--\blpage{9719}
(\byear{2020})
\end{barticle}
\endbibitem

\bibitem[\protect\citeauthoryear{Kingma}{2014}]{kingma2014adam}
\begin{botherref}
\oauthor{\bsnm{Kingma}, \binits{D.P.}}:
Adam: A method for stochastic optimization.
arXiv preprint arXiv:1412.6980
(2014)
\end{botherref}
\endbibitem

\bibitem[\protect\citeauthoryear{Li et~al.}{}]{gligen-sem}
\begin{botherref}
\oauthor{\bsnm{Li}, \binits{Y.}},
\oauthor{\bsnm{Liu}, \binits{H.}},
\oauthor{\bsnm{Wu}, \binits{Q.}},
\oauthor{\bsnm{Mu}, \binits{F.}},
\oauthor{\bsnm{Yang}, \binits{J.}},
\oauthor{\bsnm{Gao}, \binits{J.}},
\oauthor{\bsnm{Li}, \binits{C.}},
\oauthor{\bsnm{Lee}, \binits{Y.J.}}:
GLIGEN-sem.
\url{https://huggingface.co/gligen/gligen-generation-sem}
\end{botherref}
\endbibitem

\bibitem[\protect\citeauthoryear{Zhou et~al.}{2017}]{zhou2017scene}
\begin{bchapter}
\bauthor{\bsnm{Zhou}, \binits{B.}},
\bauthor{\bsnm{Zhao}, \binits{H.}},
\bauthor{\bsnm{Puig}, \binits{X.}},
\bauthor{\bsnm{Fidler}, \binits{S.}},
\bauthor{\bsnm{Barriuso}, \binits{A.}},
\bauthor{\bsnm{Torralba}, \binits{A.}}:
\bctitle{Scene parsing through ade20k dataset}.
In: \bbtitle{Proceedings of the IEEE Conference on Computer Vision and Pattern Recognition}
(\byear{2017})
\end{bchapter}
\endbibitem

\bibitem[\protect\citeauthoryear{Rombach et~al.}{}]{stable-diffusion-v1-4}
\begin{botherref}
\oauthor{\bsnm{Rombach}, \binits{R.}},
\oauthor{\bsnm{Blattmann}, \binits{A.}},
\oauthor{\bsnm{Lorenz}, \binits{D.}},
\oauthor{\bsnm{Esser}, \binits{P.}},
\oauthor{\bsnm{Ommer}, \binits{B.}}:
Stable-Diffusion-v1-4.
\url{https://huggingface.co/CompVis/stable-diffusion-v1-4}
\end{botherref}
\endbibitem

\end{thebibliography}

\end{document}